\documentclass[nohyperref]{article}

\usepackage{microtype}
\usepackage{graphicx}
\usepackage{booktabs} % for professional tables

\usepackage{hyperref}

\usepackage[accepted]{icml2023}

\usepackage{amsmath}
\usepackage{amssymb}
\usepackage{mathtools}
\usepackage{amsthm}
\usepackage{titlesec}

\usepackage[capitalize,noabbrev]{cleveref}

\usepackage{listings}
\usepackage{algorithm}
\usepackage{wrapfig}
\usepackage{booktabs}
\usepackage{physics}
\usepackage{adjustbox}
\usepackage{stfloats}
\usepackage{subcaption}

\usepackage{cleveref}
\usepackage{stfloats}

\newcommand{\eg}[0]{\emph{e.g.},~}
\newcommand{\aka}[0]{a.k.a.~}

\newcommand*{\set}[1]{\ensuremath{\mathcal{#1}}}
\newcommand{\pr}[1]{\left(#1 \right)} %for () that scale
\newcommand{\br}[1]{\left[#1 \right]} %for [] that scale
\newcommand{\cbrace}[1]{\left\{#1 \right\}} % for {} that scale
\renewcommand{\v}[1]{\ensuremath{\mathbf{#1}}} % for vectors
\newcommand{\gv}[1]{\ensuremath{\mbox{\boldmath$ #1 $}}} % for vectors of Greek letters
\newcommand{\argmin}{\operatornamewithlimits{arg\,min}}

\theoremstyle{plain}
\newtheorem{theorem}{Theorem}[section]

\newtheorem{corollary}[theorem]{Corollary}
\theoremstyle{definition}
\newtheorem{definition}[theorem]{Definition}

\theoremstyle{remark}

\usepackage[textsize=tiny]{todonotes}

\usepackage{minitoc}
\icmltitlerunning{Equivariance with Learned Canonicalization Functions}

\begin{document}

%%%%%%%%%% TOC stuff %%%%%%%%%%
\doparttoc % Tell to minitoc to generate a toc for the parts
\faketableofcontents % Run a fake tableofcontents command for the partocs

\twocolumn[
\icmltitle{Equivariance with Learned Canonicalization Functions}



\icmlsetsymbol{equal}{*}

\begin{icmlauthorlist}
\icmlauthor{Sékou-Oumar Kaba}{equal,mcgill,mila}
\icmlauthor{Arnab Kumar Mondal}{equal,mcgill,mila}
\icmlauthor{Yan Zhang}{samsung}
\icmlauthor{Yoshua Bengio}{udem,mila}
\icmlauthor{Siamak Ravanbakhsh}{mcgill,mila}
\end{icmlauthorlist}

\icmlaffiliation{mcgill}{School of Computer Science, McGill University, Montréal, Canada}
\icmlaffiliation{mila}{Mila - Quebec Artficial Intelligence Institute, Montréal, Canada}
\icmlaffiliation{samsung}{Samsung - SAIT AI Lab, Montréal, Canada}
\icmlaffiliation{udem}{DIRO, Université de Montréal, Montréal, Canada}

\icmlcorrespondingauthor{Sékou-Oumar Kaba}{kabaseko@mila.quebec}

\icmlkeywords{Machine Learning, ICML}

\vskip 0.3in
]



\printAffiliationsAndNotice{\icmlEqualContribution} % otherwise use the standard text.

\begin{abstract}
Symmetry-based neural networks often constrain the architecture in order to achieve invariance or equivariance to a group of transformations. In this paper, we propose an alternative that avoids this architectural constraint by learning to produce canonical representations of the data. These canonicalization functions can readily be plugged into non-equivariant backbone architectures. We offer explicit ways to implement them for some groups of interest. We show that this approach enjoys universality while providing interpretable insights. Our main hypothesis, supported by our empirical results, is that learning a small neural network to perform canonicalization is better than using predefined heuristics. Our experiments show that learning the canonicalization function is competitive with existing techniques for learning equivariant functions across many tasks, including image classification, $N$-body dynamics prediction, point cloud classification and part segmentation, while being faster across the board. 
\end{abstract}

\section{Introduction}
\label{sec:intro}

% \begin{figure}[htbp]
%  % Caption and label go in the first argument and the figure contents
%  % go in the second argument
% \floatconts
%   {fig:image}
%   {\caption{Add the main figure here and add captions}}
%   {\includegraphics[width=0.5\linewidth]{images/figure1.png}}
% \end{figure}

%Two sentences to explain why making a link to human cognition is useful
The problem of designing machine learning models that properly exploit the structure and symmetry of the data is becoming more important as the field is broadening its scope to more complex problems. In multiple applications, the transformations with respect to which we require a model to be \textit{invariant} or \textit{equivariant} are known and provide a strong inductive bias \citep[\eg][]{geometric_deep,bogatskiy2022symmetry, van2020mdp, mondal2020group, celledoni2021equivariant}.

\begin{figure}[ht]
\centering
\includegraphics[width=0.9\linewidth]{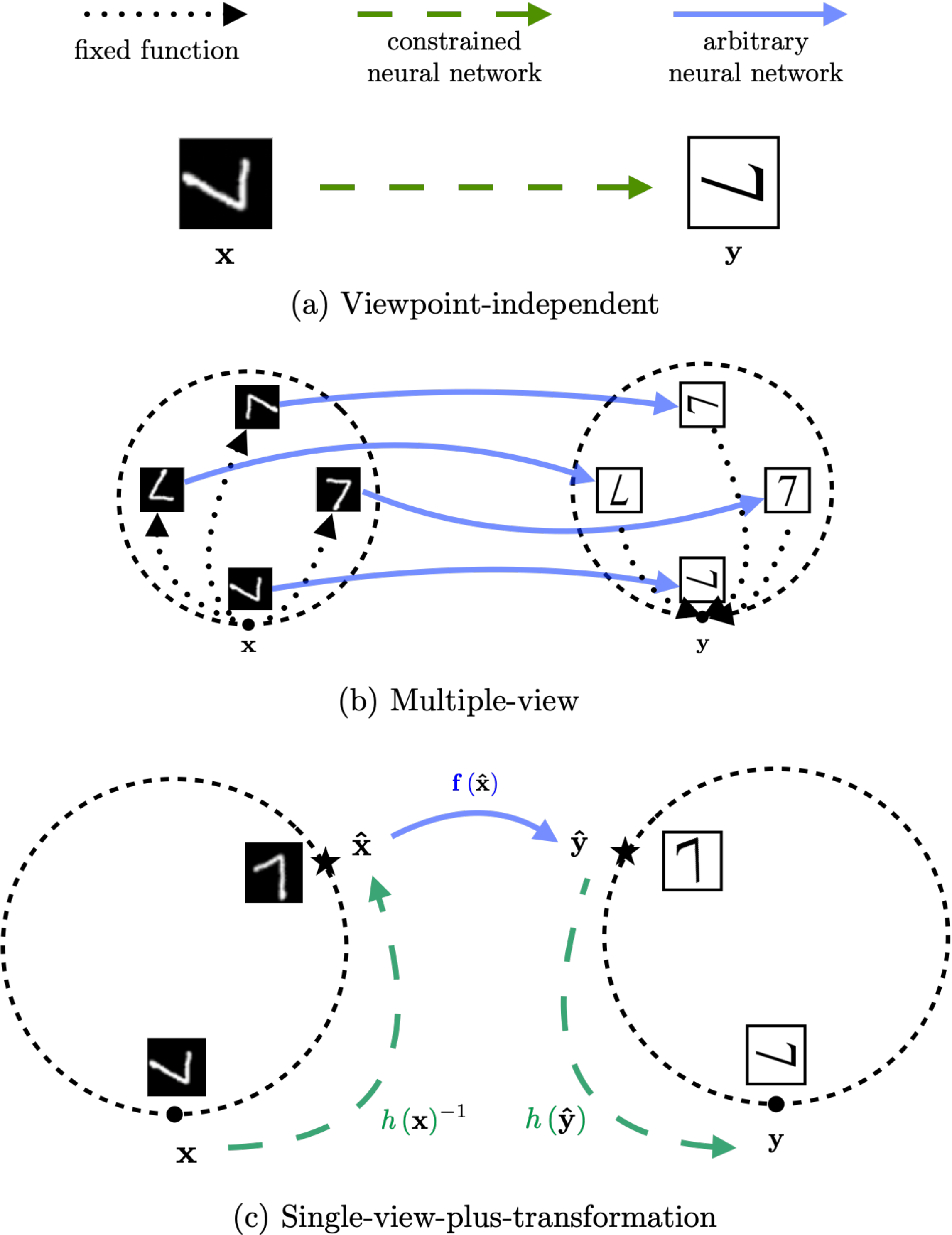}
  {\caption{\small A classification of different frameworks for equivariant predictions. In this example, the task is to restyle an MNIST digit in a rotation equivariant way. We propose a class of models that falls in the single-view-plus-transformation framework.}\label{fig:main}}
  \vspace{-6mm}
\end{figure}

As is often the case, taking a step back and drawing analogies with human cognition is fruitful here.
Human pattern recognition handles some symmetries with relative ease. When data is transformed in a way that preserves its essential characteristics, we precisely know if and how we should adapt our response.

One context in which this has been particularly well-studied in cognitive science is visual shape recognition. Experiments have shown that subjects can accurately distinguish between different orientations of an object and actual modifications to the structure of an object \citep{shepard71mentalrotation, carpenter1978}.

There are multiple ways in which this could be achieved. 
According to \citet{tarr1989}, theories of invariant shape recognition broadly fall into three categories: \textit{viewpoint-independent} models, in which object representations depend only on invariants features, \textit{multiple-view} models in which an object is represented as a set of representations corresponding to different orientations, and \textit{single-view-plus-transformation} models in which an object is converted to a canonical orientation by a transformation process.

%merge with previous paragraph
Correspondingly, similar ideas have been explored in deep learning to achieve equivariance; see \cref{fig:main}. Models that impose equivariance through constraints in the architecture \citep{symmetry_dl, cohen2016,ravanbakhsh2017equivariance} or that only use invariants as inputs \citep{villar2021scalars} can be seen as belonging to the viewpoint-independent type since the dependence of the model on symmetry transformations is trivial. The multiple-view approach includes models that symmetrize the input by averaging over all the transformations or a subset of them \citep{manay2006integral, benton2020learning,yarotsky2022universal,puny2022frame}. By contrast, the transformation approach has seen less interest. This is all the more surprising considering that evidence from cognitive science suggests that this approach is used in human visual cognition \citep{shepard71mentalrotation, carpenter1978, hinton1981frames}. When presented with a rotated version of an original pattern, the time taken by humans to do the association is proportional to the rotation angle, which is more consistent with the hypothesis that we perform a \textit{mental rotation}.
% \vspace{-1ex}
\paragraph{Present work} We introduce a systematic and general method for equivariant machine learning based on learning mappings to canonical samples. We hypothesize that among all valid canonicalization functions, some will lead to better downstream performance than others. Rather than trying to hand-engineer these functions, we may let them be learned in an end-to-end fashion with a prediction neural network. Our method can readily be used as an independent module that can be plugged into existing architectures to make them equivariant to a wide range of groups, discrete or continuous. Our approach enjoys similar expressivity advantages to methods like \textit{frame averaging} by \citet{puny2022frame}, but has several added benefits. It is simpler, more efficient, and replaces hand-engineered frames for each group by a systematic end-to-end learning approach.

Our contributions are as follows:
\vspace{-1ex}
\paragraph{Novel Framework} We introduce a general framework for equivariance to a variety of groups based on mappings to canonical samples. This framework can be plugged into any existing non-equivariant architecture.
\vspace{-1ex}
\paragraph{Theoretical Guarantees} We prove that in some settings, such models are universal approximators of equivariant functions.
\vspace{-1ex}
\paragraph{Practical Performance} We perform experiments that show that the proposed method achieves excellent results on images, physical dynamical systems and point clouds. We also support our hypothesis that learning the canonicalization function is a better strategy than designing it by hand.
\vspace{-4ex}
\paragraph{Efficient Implementations} We provide multiple variants of efficient implementations of this framework to specific domains. %Our code is available at: {\small\url{https://github.com/oumarkaba/canonical_network}}.

%The structure of this paper is as follows.

% \vspace{-1ex}
\section{Related Works}
Methods based on heuristics to standardize inputs have been around for a long time \citep{yuceer1993rotation, lowe2004distinctive}. However, these approaches require significant hand-engineering and are difficult to generalize.
An important early work is the Spatial Transformer Network \citep{jaderberg2015spatial} which learns input transformations to facilitate processing in a downstream vision task. PointNet \citep{qi2017pointnet} also proposed to learn an alignment network to encourage invariance for point cloud analysis. However, these approaches are closer to regularizers and provide no equivariance guarantees. The works of \citep{esteves2018polar,tai19a} provided equivariant versions of the Spatial Transformer using an approach based on canonical coordinates. One limitation of this approach is that it does not exactly handle equivariance to groups that are larger in dimension than the dimension of the data grid. Some recent works have proposed using learned coordinate frames for point clouds \citep{kofinas2021rototranslated,luo_2022_CVPR,du22e}. We provide theoretical and experimental evidence that the neural networks for canonicalization can be made much shallower and simpler without affecting performance. \citet{bloem2020probabilistic}, introduce the concept of \textit{representative equivariant}, which is similar to what we implement in this work. Finally, other recent works \citep{winter2022unsupervised,vadgama2022kendall} have proposed to use canonicalization in an autoencoding setup.

% \vspace{-1ex}
\section{Canonicalization Functions}
%Here, we introduce canonicalization functions for equivariance, derive some properties of this framework and instantiate it on several group.

\subsection{Problem Setting}
We are interested in learning functions $\gv{\phi}: \set{X}\to \set{Y} $ with inputs $\v{x} \in \set{X} $ and outputs $\v{y} \in \set{Y} $ belonging to finite-dimensional normed vector spaces. We will consider a set of linear symmetry transformations $T \subset \mathrm{GL} \pr{\set{X}}$, where $\mathrm{GL} \pr{\set{X}}$ is the set of invertible matrices over the vector space $\set{X}$. This is described by a \emph{group representation} $\rho: G \to T$, where $G$ is an abstract group. Without loss of generality, we can assume that $\rho$ is a group isomorphism. Therefore, the inverse $\rho^{-1}: T \to G$ is defined.
In this context, a function $\gv{\phi}$ is \emph{$G$-equivariant} if
\begin{align}
\gv{\phi}\pr{ \rho\pr{g}  \v{x}} = \rho'\pr{g} \gv{\phi}\pr{\v{x}} , \ \forall \ g, \v{x} \in G \times \set{X},
\end{align}
where the \emph{group action} $\rho$ on the input and the group action $\rho'$ on the output will be clear from the context. In particular, when $\rho'\pr{g} = I$, we say that $\gv{\phi}$ is \emph{invariant}. We use $\rho\pr{H}$ to denote the image of the subset $H$ under $\rho$.

The set $\rho\pr{G}\v{x} =  \cbrace{ \rho\pr{g} \v{x} \mid \forall \ g \in G }$ is the \emph{orbit} of the element $\v{x}$. It is the set of elements to which $\v{x}$ can be mapped by the group action. The set of orbits denoted by $\set{X}/G$ forms a partition of the set $\set{X}$.

% \vspace{-1ex}
\subsection{General Formulation}
The invariance requirement on a function $\gv{\phi}$ amounts to having all the members of a group orbit mapped to the same image by $\gv{\phi}$. It is thus possible to achieve invariance by appropriately mapping all elements to a canonical orbit representative before applying any function. For equivariance, elements can be mapped to a canonical sample and, after a function is applied, transformed back according to their original position in the orbit. This can be formalized by writing the equivariant function $\gv{\phi}$ in \textit{canonicalized form} as
\begin{align}
& \gv{\phi}\pr{\v{x}} = h'\pr{\v{x}} \v{f}\pr{h\pr{\v{x}}^{-1} \v{x} }, \label{eq:model}
\end{align}
where the function $\v{f}: \set{X}\to \set{Y}$ is called the \emph{prediction function} and the function $h: \set{X}\to \rho\pr{G}$ is called the \emph{canonicalization function}. Here $h\pr{\v{x}}^{-1}$ is the inverse of the representation matrix and $h'\pr{\v{x}} = \rho'\pr{\rho^{-1}\pr{h\pr{\v{x}}}}$ is the counterpart of $h\pr{\v{x}}$ on the output.

Equivariance in \cref{eq:model} is obtained for any prediction function if the canonicalization function is itself $G$-equivariant, $h\pr{\rho\pr{g} \v{x}} = \rho\pr{g} h\pr{\v{x}} \ \forall \ g, x \in G \times \set{X}$. 
\footnote{Symmetric inputs in $\set{X}$ pose a problem if we use the standard definition of equivariance for the canonicalization function. We explain this in Appendix \ref{apd:symmetric} and introduce the concept of relaxed equivariance that solves this issue.}

It may seem like the problem of obtaining an equivariant function has merely been transferred in this formulation. This is, however, not the case: in \cref{eq:model}, the equivariance and prediction components are effectively decoupled. The canonicalization function $h$ can therefore be chosen as a simple and inexpressive equivariant function, while the heavy-lifting is done by the prediction function $\v{f}$. 
%In Section \ref{sec:design} we introduce a way to obtain equivariant canonicalization functions for any group.

% \vspace{-1ex}
\subsection{Partial Canonicalization and Lattice of Subgroups}

%In some settings, it may still be inconvenient or difficult to satisfy the equivariance condition for the canonicalization function.
A more general condition can be formulated, such that the decoupling is partial. This enables us to impose part of the symmetry constraint on the prediction network and use canonicalization for ``additional'' symmetries. This could, for example, be used to imbue a translation equivariant architecture, like a CNN, with rotation equivariance.
\begin{theorem}
\label{th:subgroup}
For some subgroup $K\leq G$, if $\forall \ g, \v{x} \in G \times \set{X}$ there exists a $k\in K$ such that
\begin{align}
\label{eq:general}
h\pr{\rho\pr{g} \v{x}} = \rho\pr{g} h\pr{\v{x}}\rho\pr{k},
\end{align}
and the prediction function $\v{f}$ is $K$-equivariant, then $\gv{\phi}$ defined in  \cref{eq:model} is $G$-equivariant. 
\end{theorem}
The proof follows in Appendix \ref{apd:second}. This is equivalent to saying that the canonicalization function should output a coset in $G/K$ in an equivariant way, the applied transformation being chosen arbitrarily within the coset.

This can be simplified when the group factors into a semi-direct product using the following result.
\begin{theorem}
\label{th:normal}
If $K$ is a normal subgroup such that $G \simeq J \ltimes K$, condition \cref{eq:general} can be realized with a canonicalization function with image $\rho\pr{J}$, and that is $J$-equivariant and $K$-invariant.
\end{theorem}
The proof follows in Appendix \ref{apd:normal}. Going back to the example of using rotation canonicalization with a CNN, Theorem \ref{th:subgroup} says that the canonicalization function should output an element of the Euclidean group transforming equivariantly under rotations of the input. Since the translation subgroup is normal, Theorem \ref{th:normal} can be used to guarantee that the canonicalization network can always simply output a rotation.
%This arbitrariness is then resolved by the $K$-equivariance of the prediction function.
% \begin{theorem}
% \label{th:subgroup}
% For some subgroup $K\leq G$, if $\forall \ g, \v{x} \in G \times \set{X}$ there exists a $k\in K$ such that
% \begin{align}
% \label{eq:general}
% h\pr{\rho\pr{g} \v{x}} = \rho\pr{g} h\pr{\v{x}}\rho\pr{k}
% \end{align}
% and the prediction function $\v{f}$ is $K$-equivariant, then $\gv{\phi}$ defined in  \cref{eq:model} is $G$-equivariant. If $K$ is a normal subgroup such that $G \simeq J \ltimes K$, this can be realized with a canonicalization function that has image $\rho\pr{J}$, and that is $J$-equivariant and $K$-invariant.
% \end{theorem}
% The proof follows in Appendix \ref{apd:second}. This is equivalent to saying that the canonicalization function should output a representation of a coset in $G/K$ in an equivariant way, the applied transformation being chosen arbitrarily within the coset. In the example of applying rotation canonicalization 
% %This arbitrariness is then resolved by the $K$-equivariance of the prediction function.

In general, when $K=\cbrace{e}$, only the canonicalization function is constrained, which is the case described at the beginning of the section. In the image domain, this would correspond to canonicalizing with respect to the full Euclidean group and using an MLP as a prediction function. The other extreme, given by $K=G$, corresponds to transforming the input in an arbitrary way and constraining the prediction function as is usually done in equivariant architectures like Group Equivariant Convolutional Neural Networks (G-CNNs) \citep{cohen2016group}. These are, respectively, the \textit{single-view-plus-transformation} and the \textit{viewpoint-independent} implementations described in the introduction. Subgroups $\cbrace{e} < K <G$ offer intermediary options; 
the lattice of subgroups of $G$, therefore, defines a family of models. Since equivariance to a smaller group is less constraining for the prediction function, set inclusion in the subgroup lattice is equivalent to increased expressivity for the corresponding models.
%The \textit{single-view-plus-transformation} described above is recovered with $K = \cbrace{e}$ while the other extreme, the \textit{viewpoint-independent} implementation, is given by $K=G$.

% \vspace{-1ex}
\subsection{Universality Result}
\label{sec:universal}
%We now introduce a formal result supporting the claim that the decoupling between equivariance and prediction offers expressivity benefits. 
We can now introduce a more formal result on the expressivity of equivariant functions obtained with canonicalization functions. A parameterized function $\gv{\phi}$ is a universal approximator of $G$-equivariant functions if for any $G$-equivariant continuous function $\gv{\psi}$, any compact set $\set{K}\subseteq \set{X}$ and any $\epsilon > 0$, there exists a choice of parameters such that $\norm{\gv{\psi}\pr{\v{x}} - \gv{\phi}\pr{\v{x}}} < \epsilon \ \forall \ \v{x}\in \set{K} $.
\begin{theorem}
\label{th:univ}
Let $\gv{\phi}$ be a $G$-equivariant parameterized function given by \cref{eq:model} and satisfying \cref{eq:general} with $K\leq G$. Suppose that the canonicalization functions $h$ and $h'$ are continuous. Then $\gv{\phi}$ is a universal approximator of $G$-equivariant functions if the prediction function $\v{f}$ is a universal approximator of $K$-equivariant functions.
\end{theorem}
The proof follows in Appendix \ref{apd:first}.
The following corollary is especially relevant.
\begin{corollary}
A $G$-equivariant parameterized function $\gv{\phi}$ written as \cref{eq:model} with a $G$-equivariant continuous canonicalization function and a multilayer perceptron (MLP) as a prediction function is a universal approximator of $G$-equivariant functions.
\end{corollary}
This result can significantly simplify the design of universal approximators of equivariant functions since a non-universal equivariant architecture for the canonicalization function can be combined with an MLP. In particular, notice that the universality of this scheme does not hinge on the expressivity of the canonicalization network.

We note that the framework encompasses a large space of design choices, with partial canonicalization to an arbitrary subgroup $K$. It may be useful in some cases to opt for partial canonicalization and defer some equivariance to the prediction function. This reduces the number of independent parameters of the prediction function, which can help with generalization and efficiency. A useful design pattern would therefore be to find the maximal subgroup $K$ of $G$, such that a universal approximator of $K$-equivariant functions can be efficiently implemented for the prediction function. Following \cref{th:univ}, this allows universal approximation of $G$-equivariant functions.

% \vspace{-1ex}
\section{Design of Canonicalization Functions}
\label{sec:design}
\begin{figure}
\centering
\hspace{1ex}
  \begin{subfigure}[b]{0.45\columnwidth}
\centering
    \includegraphics[width=1\linewidth]{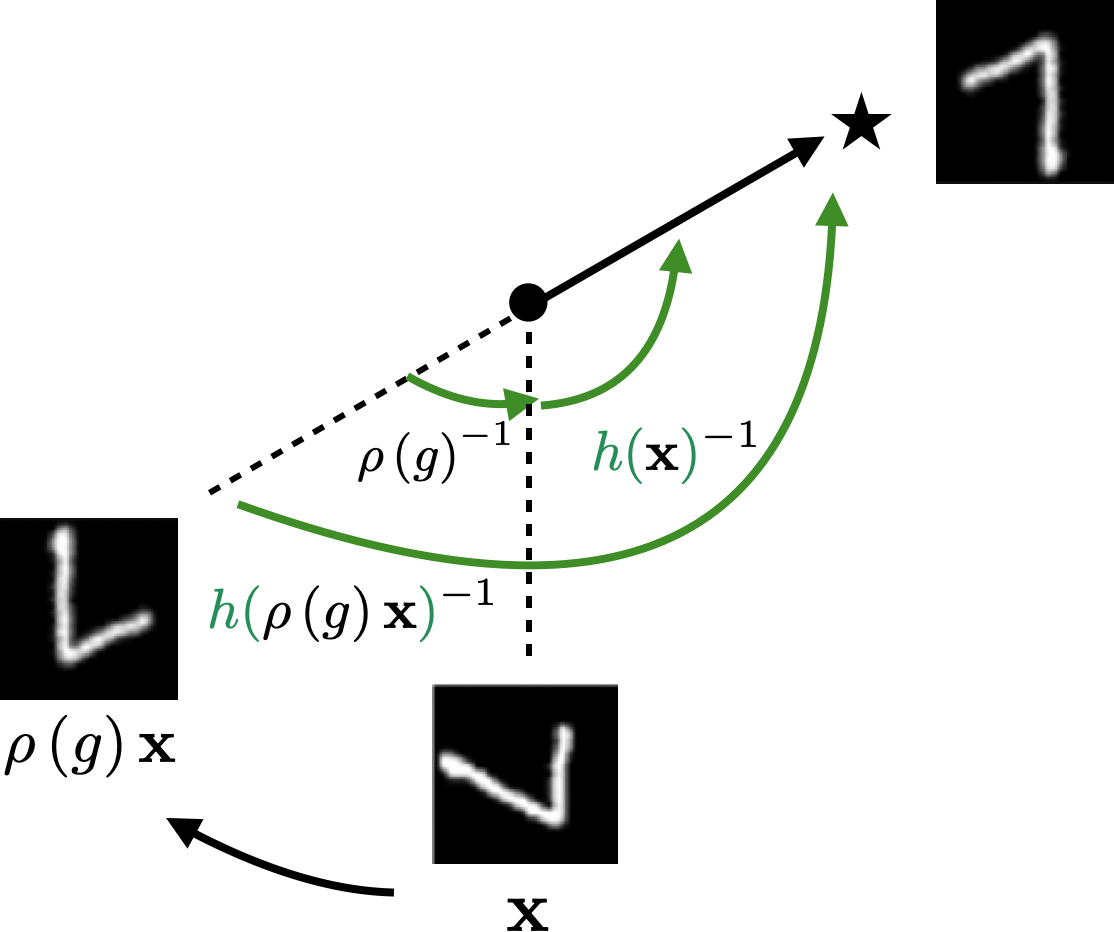}
    % \vspace{1ex}
    \caption{Direct approach}
    \label{fig:direct}
  \end{subfigure}
  \hfill %%
  \begin{subfigure}[b]{0.45\columnwidth}
\centering
    \includegraphics[width=\linewidth]{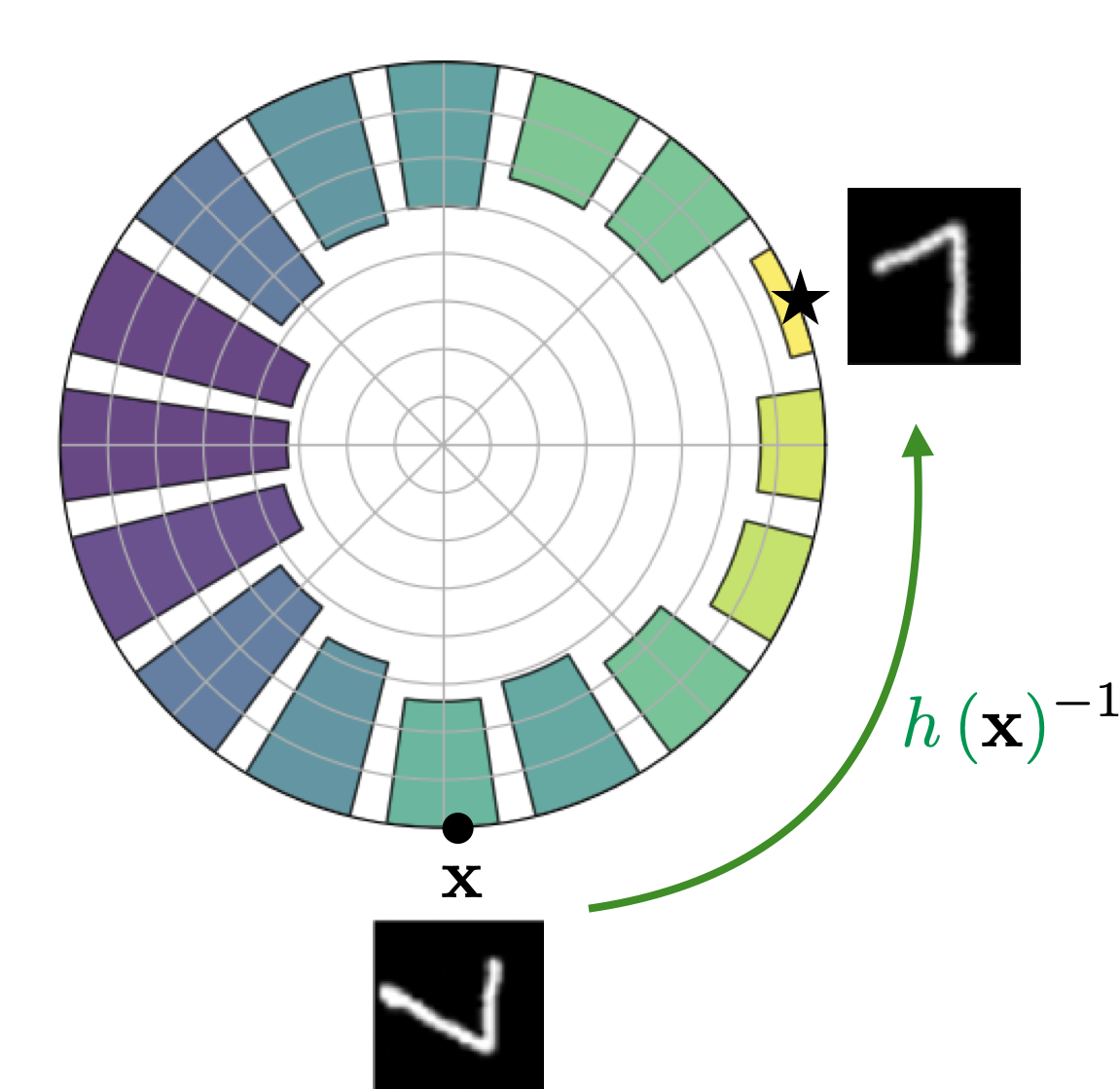}
    % \vspace{0.5ex}
    \caption{Optimization approach}
    \label{fig:optim}
  \end{subfigure}
  \caption{\small Two general approaches to canonicalization. In the direct approach, an equivariant neural network outputs the transformation. In the optimization approach, a function of the input is minimized to obtain the canonical sample.}
  \vspace{-2mm}
\end{figure}

%In this section, we elaborate on how suitable canonicalization functions can be obtained in different settings.
% \subsection{Direct Approach and Optimization Approach}
The canonicalization function can be chosen as any existing equivariant neural network architecture with the output being a group element; we call this the \textit{direct approach} (figure \ref{fig:direct}). For permutation groups and Lie groups, an equivariant multilayer perceptron \citep{symmetry_dl, finzi2021practical} can be used. We provide examples of implementations in the next section.

We also introduce an alternative method, which we call the \textit{optimization approach} (\cref{fig:optim}). The canonicalization function can be defined as
\begin{align}
\label{eq:optim}
h\pr{\v{x}} \in \argmin_{\rho\pr{g}\in \rho\pr{G}} s\pr{ \rho\pr{g}, \v{x}},
\end{align}
where $s : \rho\pr{G} \times \set{X}\to \mathbb{R}$ can be a neural network. When a set of elements minimize $s$, one is chosen arbitrarily. $s$ has to satisfy the following equivariance condition
\begin{align}
\label{eq:optim_equiv}
&s\pr{ \rho\pr{g}, \rho\pr{g_1}\v{x}} \\
&= s\pr{ \rho\pr{g_1}^{-1}\rho\pr{g}, \v{x}}, \ \forall g_1 \in G,\notag
\end{align}
and has to be such that argmin is a subset of a coset of the stabilizer of $\v{x}$. This last condition essentially means that the minimum in each orbit should be unique up to input symmetry. In \cref{apd:optim}, we prove that these are sufficient conditions to have a suitable canonicalization function.

The equivariance condition on $s$ can be satisfied using an equivariant architecture. Remarkably, it can also be satisfied using a non-equivariant function $E: \set{X}\to \mathbb{R}$ and defining
\begin{align}
s\pr{ \rho\pr{g}, \v{x}} = E\pr{ \rho\pr{g}^{-1} \v{x}}. \label{eqn:simple optim}
\end{align}
We will call the function $E$ energy. Intuitively, $E$ represents a distance between the input and the canonical sample of the orbit and is therefore minimized when $\rho\pr{g}$ is the transformation that maps to the canonical sample.
% For continuous groups, a function with unique minimum can be learned with Input Convex Neural Networks (ICNN) \citep{amos2017}, but satisfying this condition for each orbit is more challenging. The minimization in equation \ref{eq:optim} can then be performed using any standard convex optimization method. 
% For practical application, we are typically interested in low-dimensional Lie groups, which makes second-order optimization methods usable. The gradient of the neural network with respect to the parameters can be obtained efficiently (e.g. without having to differentiate through all the optimization steps) with implicit differentiation \citep{blondel2022efficient}. 
%In some cases, it is also possible to conveniently perform the optimization for permutation groups using an optimal transport formulation \citep{mena2018learning, blondel2018smooth, cuturi2019}.

This implementation presents a close analogy with the mental rotation phenomenon described in the introduction, as humans try to minimize the distance between their representation of an object and the canonical one. As such, it is expected that the optimization process will take more iterations when the input sample is farther away in orbit from the canonical sample. This is consistent with the experimental evidence for mental rotation \citep{shepard71mentalrotation, carpenter1978}.

Simultaneous minimization and learning of $s$ results in a bi-level optimization problem \cite{gould2016differentiating, liu2021investigating}. This can be performed in a variety of ways, including using implicit methods \cite{blondel2022efficient}.
Next, we elaborate on how suitable canonicalization functions can be obtained in different settings.

% \vspace{-1ex}
\subsection{Euclidean Group}
\label{sec:euclidean}
% \subsubsection{Euclidean Group}
The Euclidean group $E\pr{d}$ describes rotation, translation, and reflection symmetry. Domains in which this type of symmetry is especially relevant include computer vision,  point cloud modelling and physics applications. Below we give design principles to obtain equivariant models for image and point cloud inputs.
% \paragraph{Image Input.}
% In order to build networks equivariant to $E(2)$, we need a canonicalization function that outputs an element of $E(2)$ given an image. This can be achieved by using a G-CNN \citep{cohen2016}. We explain this by taking an example of building a canonicalization network for the Special Euclidean  $SE(2)$ group. For an image given by a $2$D signal $I:\mathbb{R}^2 \to \mathbb{R}$, $SE(2)$ action consists of translation by $t \in \mathbb{R}^2$, and rotation by a $2\times 2$ rotation matrix -- that is $\rho(g) = (t, r)$ and 
% $$[\rho(g) \cdot I](x) = I(r^{-1}(x-t)) \quad \forall x\in \mathbb{R}^2,$$  
% where $x$ is pixel position. An ideal canonicalization network should map images to elements of the $SE(2)$ group $h: I \mapsto (t_I,r_I)$ such that 
% \begin{align}\label{eq:se2}
% h((t, r) \cdot I)=(t, r) \cdot h(I)= (t+r t_I,r r_I)
% \end{align}
% for any SE transformation $(t, r)$. 

% \vspace{-1ex}
\paragraph{Image Input.}
Elements of the Euclidean group can be written as $\pr{\v{O}, \v{t}}$, where $\v{O} \in \mathbb{R}^{n\times n} $ is an orthogonal matrix and $\v{t}\in \mathbb{R}^{n}$ is an arbitrary translation vector. We consider the space of image inputs $I \in \mathcal{X}$ as given by a $2$ dimensional signal $I:\mathbb{R}^2 \to \mathbb{R}^C$, where $C$ is the number of input channels. We adopt a continuous description to facilitate exposition, but in practice, all the operations are discretized using interpolation \citep{eriba2019kornia}. This thus reduces to the $pnm$ group, which is the group of $n$-fold discrete rotations, reflections and discrete translations. The action of the representation on image inputs is defined by the following linear operator
$$[\rho\pr{\v{O}, \v{t}} \cdot I](\v{p}) = I(\v{O}^{-1}(\v{p}-\v{t})), \quad \forall \v{p}\in \mathbb{R}^2,$$
where $\v{p}$ is pixel position.
The canonicalization function should output an element of the $E\pr{2}$. It should also be $E\pr{2}$-equivariant, such that $h\pr{\rho\pr{\v{O}, \v{t}} \cdot {I}} = {\rho\pr{\v{O}, \v{t}}} \cdot h\pr{I}$.

This condition can be satisfied by using a 
% $E\pr{2}$-Equivariant Steerable Network \cite{weiler2019general}
Group Equivariant CNN (G-CNN) \citep{cohen2016} and the optimization approach described above.
To do this, we define the function to be optimized as $s: O\pr{2} \times \mathbb{R}^2 \times \mathcal{X} \to \mathbb{R}$. This can be reinterpreted as $s: \mathcal{X} \to \mathbb{R}^{ O\pr{2} \times \mathbb{R}^2} $, which means where the first dimension, \aka the \emph{fiber}, encodes rotation angles and $\mathbb{R}^2$ is associated with pixel positions. If $s$ is a G-CNN, it correctly satisfies the condition \cref{eq:optim_equiv}, as image rotations act on the fiber and Euclidean transformations on the pixel positions. The canonicalization is then obtained by taking the $\argmin$ over pixel positions and fibers
\begin{align}
h\pr{\v{x}} \in \argmin_{\pr{\v{O}, \v{t}}\in E(2)} s\pr{\v{x}}_{\pr{\v{O}, \v{t}}}.
\end{align}
% \footnote{ More concretely, 
% the action of the $\rho(g) =(t, r)$ is given by $$[\rho(g) \cdot f \circ I](x, \theta) = f\circ I(r^{-1}(x-t), (\theta_{r^{-1}} + \theta) \% (2\pi)),$$ 
% where $\theta_{r^{-1}}$ denotes angle corresponding to $r^{-1}$ and $\%$ denotes the remainder operation.}, 
This approach can be further simplified if we use a translation equivariant prediction network, such as a CNN-based architecture. As the translation group $T(2)$ is a normal subgroup of the Euclidean group $E(2)$, using \cref{th:normal}, we only require the canonicalization function to be equivariant to $O(2)$. This means we can average over the spatial dimension in the output feature map of the canonicalization function and only need to take an $\argmin$ along the rotation fiber dimension to identify the orientation of the image.

There are two potential problems with this approach. First, extending G-CNNs to a higher number of finer discrete rotations is computationally expensive, and it leads to artifacts. 
% due to the finer rotation of filters. 
Second, we cannot backpropagate through the canonicalization function as the $\arg\min$ operation is not differentiable. 

We can avoid the first problem by using a shallower network with a larger filter size. We empirically show why this is a sound choice for canonicalization function in  
\cref{sec:experiments}. We use the straight-through gradient estimator \citep{bengio2013estimating} to solve the second problem. Appendix \ref{apd:third} contains a \textsc{PyTorch} code snippet to perform the canonicalization function of images in a differentiable way using a G-CNN. 
% Similarly, the output of the canonicalization function can be used to invert the feature maps from the prediction network back to their original orientation in a differentiable way.

% Since CNNs are universal approximators of $T(2)$-equivariant functions \citep{yarotsky2022universal}, it follows from \cref{th:univ} that a CNN augmented with an $O(2)$ equivariant canonicalization function is a universal approximator of $E(2)$-equivariant functions.

% \vspace{-1ex}
\paragraph{Point Cloud Input.}
%Elements of the Euclidean group can be written as $\pr{\v{O}, \v{t}}$, where $\v{O} \in \mathbb{R}^{n\times n} $ is an orthogonal matrix and $\v{t}\in \mathbb{R}^{n}$ is an arbitrary translation vector.
The $n+1$ dimensional representation of the Euclidean group (defined by concatenating a constant 1 to the original vectors) is defined in the following way
\begin{align}
& \rho{\pr{\v{O}, \v{t}}} = 
\begin{pmatrix}
\v{O} & \v{t}\\
\v{t}^T & 1
\end{pmatrix}.
\end{align}
% SHow that This representation is bounded 
We seek to define an $E\pr{d}$-equivariant canonicalization function for point clouds. This can be done by defining it as $h\pr{\v{x}} = \rho{\pr{h^{O}\pr{\v{x}}, h^{t}\pr{\v{x}}}}$, where the function $h^{O}: \set{X} \to \mathbb{R}^{n\times n}$ outputs the rotation and reflection and $h^{t}: \set{X} \to \mathbb{R}^{n}$ the translation. Since the product of elements of $E\pr{n}$ is given by $\pr{\v{O}_1, \v{t}_1}\pr{\v{O}_2, \v{t}_2} = \pr{\v{O}_1\v{O}_2, \v{O}_2\v{t}_1+\v{t}_2}$, the equivariance condition requires that we have
\begin{align}
% \rho{\pr{h^{O}\pr{\rho{\pr{\v{O}, \v{t}}} \v{x}}, h^{t}\pr{\rho{\pr{\v{O}, \v{t}}} \v{x}}}} &= \rho{\pr{\v{O}, \v{t}}} \rho{\pr{h^{O}\pr{\v{x}}, h^{t}\pr{\v{x}}}}\\
%\pr{h^{O}\pr{\rho{\pr{\v{O}, \v{t}}} \v{x}}, h^{t}\pr{\rho{\pr{\v{O}, \v{t}}} \v{x}}} &= \pr{\v{O}, \v{t}} \pr{h^{O}\pr{\v{x}}, h^{t}\pr{\v{x}}}\\
% &\pr{h^{O}\pr{\rho{\pr{\v{O}, \v{t}}} \v{x}}, h^{t}\pr{\rho{\pr{\v{O}, \v{t}}} \v{x}}} \\
% &= \pr{\v{O} h^{O}\pr{\v{x}}, \v{O} h^{t}\pr{\v{x}} + \v{t}}\notag
&{h^{O}\pr{\rho{\pr{\v{O}, \v{t}}} \v{x}}} = \v{O} h^{O}\pr{\v{x}}, \\
&h^{t}\pr{\rho{\pr{\v{O}, \v{t}}} \v{x}} = {\v{O} h^{t}\pr{\v{x}} + \v{t}}.
\end{align}
This means that $h^{O}$ must be $O\pr{d}$-equivariant and translation invariant, and that $h^{t}$ must be $E\pr{d}$-equivariant. These constraints can be satisfied by using already existing equivariant architectures. Since most of the work will be done by a prediction function that can be very expressive, like Pointnet \citep{qi2017pointnet}, a simple and efficient architecture can be used for the canonicalization function, for example, Vector Neurons \citep{deng2021vector}. The output of $h^{O}$ can be made an orthogonal matrix by having it output $n$ vectors and orthonormalizing them with the Gram-Schmidt procedure, which is itself equivariant (Appendix \ref{apd:gram}).

Using Deep Sets \citep{deep_sets} as a backbone architecture would result in a universal approximator of $E\pr{d}$ and permutation equivariant functions, following \cref{th:univ} and Theorem 1 of \citep{segol2020On}.

\vspace{-1ex}
\subsection{Symmetric Group}
The symmetric group $S_n$ over a finite set of $n$ elements contains all the permutations of that set. This group captures the inductive bias that input order should not matter. Domains for which $S_n$-equivariance is desirable include object modelling and detection, graph representation learning, and applications in language modelling.

$S_n$-equivariant canonicalization functions can be obtained with a direct approach using existing optimal transport solvers \citep{villani2009optimal}.
For example, the Sinkhorn algorithm \cite{sinkhorn, mena2018learning} solves the entropy-regularized optimal transport problem \citep{cuturi2013sinkhorn}, which results in convex combinations of permutations (doubly-stochastic matrices) that are equivariant. In practice, this is an example of the implementation of canonicalization with the direct approach.
% However, these approaches only learn relaxations of permutations and not actual elements of the permutation group, so care needs to be taken when inverting them.
%In addition, they offer no clear way to handle sets of different sizes.
Obtaining a permutation can also be framed as an optimization problem, which makes our optimization approach in \cref{eq:optim} relevant; problems like sorting \citep{blondel20a} and optimal transport \citep{blondel2018smooth} are often formulated like this, which shows that this is a powerful paradigm.

% An alternative for $S_n$-equivariance is to consider partial canonicalization following Theorem \ref{th:subgroup}. Instead of producing a permutation of the input, the canonicalization function can instead be designed to obtain a clustering of the input, with a function $c: \set{X}\to \cbrace{1, \dots, k}$, where $k$ is the number of possible clusters. This can be made differentiable using the straight-through estimator described for the images above. Then, a corresponding permutation is obtained by sorting elements by cluster index, with elements belonging to the same cluster being sorted arbitrarily. The canonicalization function would therefore respect equation \ref{eq:general} with respect to the subgroup $K = S_{l_1}\times \dots \times S_{l_k}$, where $l_i$ is the number of elements assigned to cluster $i$, the product of symmetric groups within clusters. Heuristically, this enforces the inductive bias that set elements should be mapped to high-level classes, across which interactions will be evaluated in an expressive way. This is close in spirit to what is done in the ClusterFormer \citep{wang2022clusterformer} and Reformer \citep{kitaev2020reformer} transformer architectures.

% The prediction function $\v{f}$ would then have to be $S_{l_1}\times \dots \times S_{l_k}$ equivariant to obtain permutation equivariance. Such a function can readily be obtained with an equivariant multilayer perceptron, as derived by \citet{hierarchical}. 

% \newpage

\vspace{-1ex}
\section{Experiments}
\label{sec:experiments}
%For all experiments, training was performed using the Adam optimizer \citep{adam}, etc.
\subsection{Image classification}
% \subsubsection{Experimental setup}
We first perform an empirical analysis of the proposed framework in the image domain. We selected the Rotated MNIST dataset \citep{larochelle2007}, often used as a benchmark for equivariant architectures. The task is to classify randomly rotated digits. In \cref{tab:image:cls_comp}, we compare our method with different CNN and G-CNN \citep{cohen2016} baselines. We denote 
% the group of $n$ discrete rotations as $pn$, and 
the networks equivariant to $pn$ by putting it with the network's name (e.g. G-CNN(p4)). The training and architecture details are provided in \cref{appndx:image:exp}. For the canonicalization function, we choose a shallow G-CNN with three layers. The first layer is a lifting layer which maps the signal in the pixel space to the group with filters that are the same size as the input image.
This is followed by ReLU nonlinearity and group equivariant layers with $1\times 1$ filters.

%Having a larger filter reduces artifacts when we extend G-CNNs to higher order rotations which we show in Section \ref{subsubsec:ablation} contributes significantly to the performance improvement.

\begin{table}%{r}{.9\linewidth}
\centering
\small
\caption{\small Comparison with the existing work for Rotated-MNIST.}
\label{tab:image:cls_comp}
\label{tab:mnist_results}
\begin{adjustbox}{width=0.6\linewidth}
\begin{tabular}{lll}
  \toprule
  \bfseries Method & \bfseries Error \% $\downarrow$ \\
  \midrule
  CNN (base) & 4.90 $\pm$ 0.20\\
  G-CNN (p4) & 2.28 $\pm$ 0.00\\
  G-CNN (p4 \& = params) & 2.36 $\pm$ 0.15\\
  G-CNN (p64 \& = params) & 2.28 $\pm$ 0.10\\
  %CNN (= params) & 4.80 $\pm$ 0.37\\
  \midrule
\multicolumn{2}{c}{Ours} \\
\midrule
  CN(PCA)-CNN & 3.35 $\pm$  0.21\\
  CN(p4 \& frozen)-CNN & 3.91 $\pm$ 0.12\\
  \midrule
  CN(OPT)-CNN & 3.35 $\pm$ 0.00\\
  CN(p4)-CNN & 2.41 $\pm$ 0.10\\
  CN(p64)-CNN & 1.99 $\pm$ 0.10\\
  %Ours (p4m) & 6.11 $\pm$ 0.8\\
  \bottomrule
  \end{tabular}

\end{adjustbox}
\vspace{-2mm}
\end{table}

We learn the canonicalization function end-to-end with a CNN as the prediction function (CN($pn$)-CNN). We also implement \cref{eqn:simple optim} as CN(OPT)-CNN. Our $E$ converts the input image into a point cloud representation, which is fed into a PointNet that produces the energy. We use gradient descent to optimize this energy with respect to the input rotation for a small number of steps. This procedure is visualized in \cref{fig:optim}.

For the pure G-CNN-based baseline, we provide the value reported by \citet{cohen2016} and design a variant which has similar architecture to CNN (base) while matching the number of parameters of our CN(p4)-CNN. We call this G-CNN (p4 \& = params). 

Lastly, we consider variants where the canonicalization function is not learned. The first one is a G-CNN similar to CN($pn$) but with weights frozen at initialization. We call them CN(p4 \& frozen)-CNN and CN(p64 \& frozen)-CNN. For the second one, canonicalization is performed by finding the orientation of the digits using Principal Component Analysis (PCA) and we refer to it as CN(PCA)-CNN.

\vspace{-1ex}
\paragraph{Results}

As reported in \cref{tab:image:cls_comp} the direct canonicalization approach outperforms the CNN-based baseline and is comparable to G-CNNs. The optimization version does not perform as well, even if it is still better than the non-equivariant baseline. We have found that this is because gradient descent can get stuck in flat regions. We see that using a fixed canonicalization function technique like PCA or canonicalization function with frozen parameters improves performance over the CNN baseline. Learning the canonicalization function provides a significant performance improvement. 
%We further notice that adding a canonicalization function for reflections hurts performance. This is, in fact, not surprising, as reflection augmentations were not used to build the rotated-MNIST dataset. Adding reflection invariance while the data does not possess it may introduce ambiguity between certain digits like 2 and 5. 

\begin{table}%{r}{.9\linewidth}
\centering
\small
\caption{\small Ablation study on the effect of augmentation.}
\label{tab:mnist_ablation}
\begin{adjustbox}{width=0.6\linewidth}
\begin{tabular}{lll}
  \toprule
  \bfseries Method & \bfseries Error \% $\downarrow$ \\
  \midrule
  CNN (base) & 4.90 $\pm$ 0.20\\
\midrule
  CNN (rotation aug.) & 3.30 $\pm$ 0.20\\
  CN(pretrained)-CNN & 2.05 $\pm$ 0.15\\
  CN(p64)-CNN & 1.99 $\pm$ 0.10\\
  \bottomrule
  \end{tabular}

\end{adjustbox}
\vspace{-4mm}
\end{table}

\begin{figure}[ht!]%{r}{0.5\linewidth}
  \centering
   \vspace{1em}
   \includegraphics[width=0.75\linewidth, trim={1cm 0.5cm 1cm 1cm}]{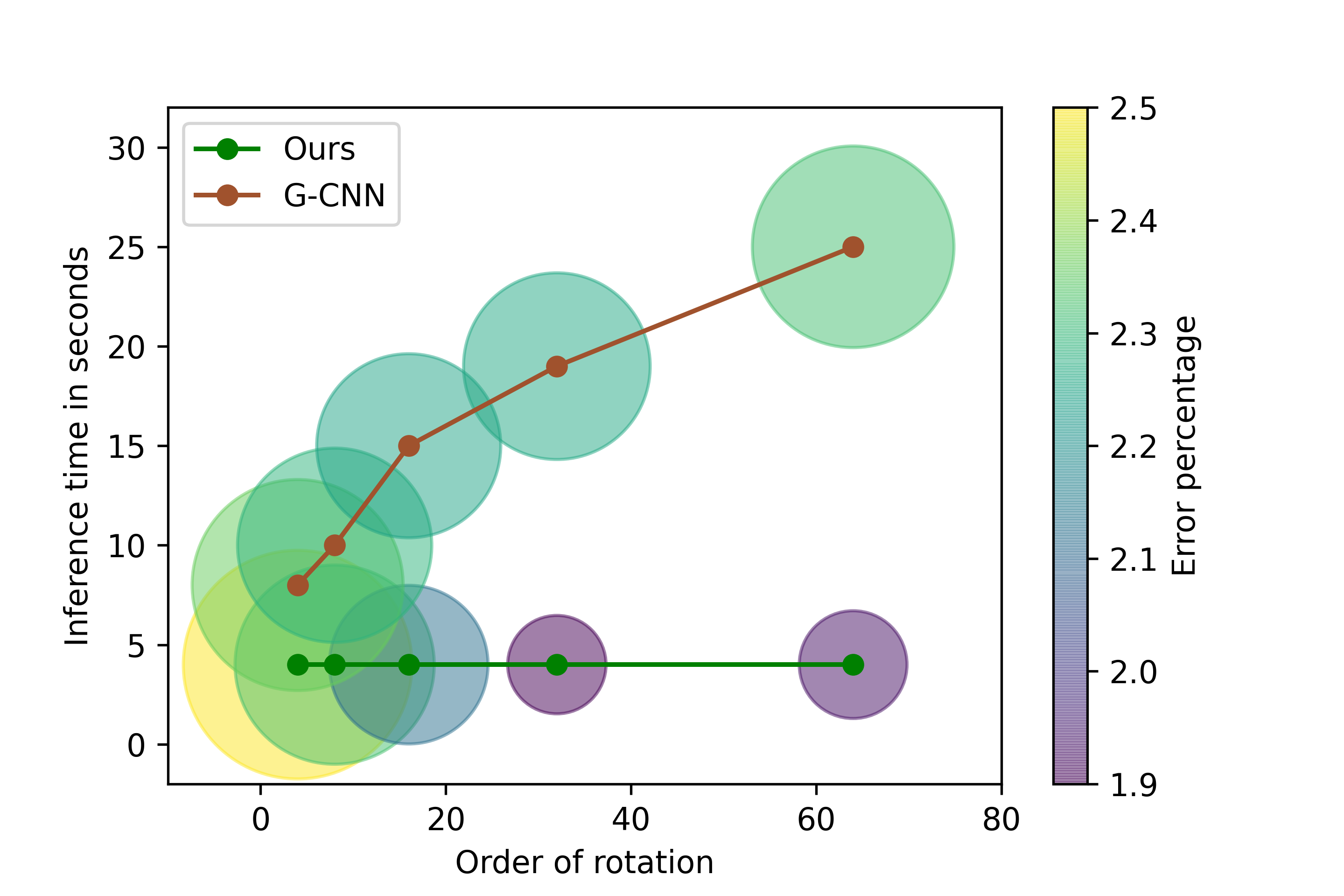}
\caption{\small Inference time comparison of our method with G-CNN with increasing order of rotations.}
  \label{fig:inference_speed}
  \vspace{-2mm}
\end{figure}

\vspace{-1ex}
\paragraph{Ablation study}
We further seek to understand if learning the canonicalization performs better mainly because a meaningful function is learned, or because this implicitly augments the prediction CNN with rotations during training. We perform an ablation study to investigate this.
First, we compare with a CNN trained with random rotation augmentations. Second, we compare with a setup we call CN(pretrained), in which canonicalization is learned along with a CNN prediction network. Then the prediction function is reinitialized and trained from scratch while the canonicalization function is fixed. If a meaningful canonicalization is learned, this setup should perform close to the one where the canonicalization is learned end-to-end. We see from the results of \cref{tab:mnist_ablation} that this is the case. The pretrained canonicalization performs almost as close as the end-to-end one and significantly better than data augmentation. We also visualize the canonicalized samples for different canonicalization techniques and order of rotation in \cref{appdx:image:add_results}, confirming that meaningful canonicalization is learned.
% Next, we perform experiments to understand the role of different components in our model using the group of $n$ discrete rotations $(pn)$. 

\vspace{-1ex}
\paragraph{Inference time}
Next, we compare the inference time of our model with pure G-CNN-based architectures. For this experiment, we take the CNN architecture of our predictor network and replace the convolutions with group convolutions. As increasing the rotation order in G-CNN requires more copies of rotated filters in the lifting layer and more parameters in the subsequent group convolution layers, we decreased the number of channels to keep the number of parameters the same as our model. Figure \ref{fig:inference_speed} shows that although G-CNN's performance is slightly better for the p4 group, increasing the order of discrete rotations improves our model's performance significantly compared to G-CNN. In addition to performance gain, our model's inference speed remains more or less constant while encoding invariance to higher-order rotations due to the shallow canonicalization network. This makes our approach more suitable for building equivariance for bigger groups and network architectures.

% \begin{table}[hbtp]
% \floatconts
%   {tab:example-booktabs}
%   {\caption{Ablation study on the impact learning the canonicalization function}}
%   {\begin{tabular}{l|lllllll}
%   \toprule
%   \bfseries Model type & \multicolumn{4}{c}{ \bfseries Order of the discrete rotation group} \\
%     & p4 & p8 & p16 & p32 & p64& \\
%   \midrule
%   w gradients & 2.52 $\pm$ 0.12& 2.37 $\pm$ 0.09& 2.2 $\pm$ 0.08 & 2.05 $\pm$ 0.15 & 2.01 $\pm$ 0.09\\
%   w/o gradients & 4.31 $\pm$ 0.15& 4.12 $\pm$ 0.09& 3.9 $\pm$ 0.14 & 3.85 $\pm$ 0.18 & 3.67 $\pm$ 0.15\\
%   \bottomrule
%   \end{tabular}}
% \end{table}

% \subsection{CIFAR-10}

% \subsection{Discussion}
% \paragraph{Adding the canonicalization function is beneficial if there is symmetry in the domain}

% \paragraph{Learning the canonicalization function improves performance}

% \paragraph{Performance peaks with increased expressivity of the canonical network}

% \begin{table}[]
%     \centering
%     \begin{tabular}{lllll}
%   \toprule
%   \bfseries Method & Architecture & \bfseries Error \% \downarrow \\
%   \midrule
%   CNN & ResNet44 & 9.45 \\
%   GroupCNN (p4m) & ResNet44 & 6.46\\
%   Ours (p1m) & ResNet44 & 6.9 $\pm$ 0.5\\
%   Ours (p4m) & ResNet44 & 9.9 $\pm$ 0.5\\
%   % CNN  & ResNet18 & 8.9 $\pm$ 0.2\\
%   % Ours (p1m) & ResNet18 & 7.6 $\pm$ 0.1\\
%   % Ours (p4m) & ResNet18 & 9.4 $\pm$ 0.3\\
%   \bottomrule
%   \end{tabular}
% \caption{Performance on CIFAR10}
%     \label{tab:my_label}
% \end{table}

%\vspace{-1ex}

% \vspace{-1ex}
\subsection{$N$-body dynamics prediction}
% \subsubsection{Experimental setup}
Simulation of physical dynamics is an important class of $E(3)$-equivariant problems due to the symmetry of physical laws under rotations and translations. We evaluate our framework in this setting with the $N$-body dynamics prediction task proposed by \cite{kipf2018neural} and \cite{fuchs2020se3transformers}. In this task, the model has to predict the future positions of 5 charged particles interacting with Coulomb force given initial positions and velocities. We use the same version of the dataset and setup as \cite{satorras2021n}.

\begin{table}[t] 
\small
  \centering
  \caption{\small Test MSE for the N-body dynamics prediction task.}
  \label{table:n_body}
  \begin{tabular}{lc}
  \toprule
    Method & MSE  \\
    \midrule
    Linear & 0.0819  \\
       SE(3) Transformer & 0.0244 \\ 
       TFN & 0.0155  \\ 
        GNN &  0.0107 \\ 
       Radial Field & 0.0104 \\
        {EGNN} & {0.0071} \\ 
    % \midrule
        {FA-GNN} & {0.0057} \\ 
    \midrule
        % GNN (Ours) &  0.0107 & -  \\ 
        \textbf{CN-GNN} & \textbf{0.0043} ${\pm}$ \textbf{0.0001}  \\ 
        {CN-GNN-$O(3)$} & {0.0045} $\pm$ 0.0001 \\ 
        {CN-GNN} (frozen) & {0.0085} $\pm$ 0.0002 \\ 
    \bottomrule
  \end{tabular}
  \vspace{-2mm}
\end{table}

For this experiment, our architecture uses a simple 2-layer Vector Neurons version of the Deep Sets architecture for the canonicalization function \cite{deng2021vector,deep_sets}. The prediction function is a 4-layer Graph Neural Network (GNN) with the same hyperparameters as the one used in \cite{satorras2021n}, and \cite{puny2022frame} for a fair comparison. The architecture of the prediction network was, therefore, not optimized. The canonicalization network is much smaller than the prediction GNN, with around 20 times fewer parameters. This allows us to test the hypothesis again that only a simple canonicalization function is necessary to achieve good performance.
% We give more details on the architecture and training setup in Appendix \ref{apd:n-body}.
\cref{apd:n-body} contains more details on the architecture and training setup.

\vspace{-1ex}
\paragraph{Results}
\cref{table:n_body} shows that we obtain state-of-the-art results.
The improvement with respect to Frame Averaging is significant, showing that learning the canonicalization provides an important advantage. Our approach also does better than all the intrinsically equivariant (or \textit{viewpoint-independent}) baselines both in accuracy and efficiency. This shows that canonicalization can be used to obtain equivariant models with high generalization abilities without sophisticated architectural choices.

\vspace{-1ex}
\paragraph{Ablation study}

We also test variants of the model. First, we test on a variant of the model where the canonicalization is only learned for the $O\pr{3}$ part of the transformation and where the translation part is given by the centroid. Since, for this system, all the masses are identical, this is the same as the center of mass of the system. The result is reported in \cref{table:n_body} as {CN-GNN-$O(3)$}. We obtain only marginally worse performance compared to the fully trained canonicalization function. This shows that, in this setting, the centroid provides an already suitable canonicalization function, which is expected given the physical soundness of choosing the center of mass as the origin of the reference frame. Since the learned translation canonicalization performs on par with this physically motivated canonicalization, this also validates the method.

The comparison with Frame Averaging is also insightful. PCA-based Frame Averaging can also be motivated from a physical point of view since this method is equivalent to identifying the principal axes using the tensor of inertia. It is, therefore, a physical heuristic for $O\pr{3}$ canonicalization. By contrast with the translation canonicalization with the centroid, for orthogonal transformations learning, the canonicalization performs significantly better.

Second, we compare with a version of the model where the weights of the canonicalization function are frozen at initialization. This canonicalization still provides $E(n)$-equivariance and, as expected, provides a significant improvement of more than 20\% with respect to the GNN prediction function alone. However, the learned canonicalization function provides a close to 50\% improvement in performance compared to this fixed canonicalization.

% - O(3) canonicalization + centrer of mass
% - T(3) canonicalization + principal axes
% - center of mass + principal axes
% - frozen weights

% \vspace{-1ex}
\subsection{Point cloud classification and segmentation}

% \subsubsection{Experimental setup}

We use the ModelNet40 \cite{wu20153d} and ShapeNet \cite{shapenet2015} datasets for experiments on point clouds. The ModelNet40 dataset consists of 40 classes of 3D models, with a total of 12,311 models. 9,843 models were used for training, and the remaining models were used for testing in the classification task. The ShapeNet dataset was used for part segmentation with the ShapeNet-part subset, which includes 16 categories of objects and more than 30,000 models. In the classification and segmentation task, the train/test rotation setup adhered to the conventions established by \cite{esteves2018learning} and adopted by \cite{deng2021vector}. Three settings were implemented: $z/z$, $z/SO(3)$, and $SO(3)/SO(3)$. The notation $z$ denotes data augmentation with rotations around the z-axis during training, while $SO(3)$ represents arbitrary rotations. The notation $x/y$ denotes that transformation $x$ is applied during training and transformation $y$ is applied during testing.

\begin{table}[ht]
\small
  \centering
  \caption{\small Test classification accuracy of different point cloud models on the ModelNet40 dataset \cite{wu20153d} in three train/test scenarios. This table is borrowed from \cite{deng2021vector}. $z$ here stands for aligned data augmented by random rotations around the vertical axis, and $\mathrm{SO}(3)$ indicates data augmented by random 3D rotations.}
  \label{tab:exp:cls}
\resizebox{\columnwidth}{!}{
  \begin{tabular}{lccc}
    \toprule
    Method & $z/z$ & $z/\mathrm{SO}(3)$ & $\mathrm{SO}(3)/\mathrm{SO}(3)$ \\
    \midrule
    \multicolumn{4}{c}{Point / mesh inputs} \\
    \midrule
    PointNet \cite{qi2017pointnet} & 85.9 & 19.6 & 74.7 \\
    DGCNN \cite{wang2019dynamic} & 90.3 & 33.8 & 88.6 \\
    % VN-DGCNN (w/o BN) & 88.9 & 88.9 & 88.6 \\
    VN-PointNet & 77.5 & 77.5 & 77.2 \\
    VN-DGCNN & 89.5 & 89.5 & 90.2 \\
    \midrule
    % --------------------
    % input: multiview & sphere projection
    % --------------------
    %EMVnet \cite{esteves2019equivariant} & \textbf{94.4} & - & \textbf{91.1} \\
    % $a^3$S-CNN \cite{liu2018deep} & 89.6 & 87.9 & 88.7 \\
    % --------------------
    % input: point
    % --------------------
    PCNN \cite{atzmon2018point} & 92.3 & 11.9 & 85.1  \\
    ShellNet \cite{zhang2019shellnet} & 93.1 & 19.9 & 87.8 \\
    PointNet++ \cite{qi2017pointnetpp} & 91.8  & 28.4 & 85.0 \\
    PointCNN \cite{li2018pointcnn} & 92.5 & 41.2 & 84.5 \\
    % 
    % --------------------
    % input: mesh or point
    % --------------------
    Spherical-CNN \cite{esteves2018learning} & 88.9 & 76.7 & 86.9 \\
    $a^3$S-CNN \cite{liu2018deep} & 89.6 & 87.9 & 88.7 \\
    % --------------------
    % input: robust point
    % --------------------
    \midrule
    SFCNN \cite{rao2019spherical} & 91.4 & 84.8 & 90.1 \\
    TFN \cite{thomas2018tensor} & 88.5 & 85.3 & 87.6 \\
    RI-Conv \cite{zhang2019rotation} & 86.5 & 86.4 & 86.4 \\
    SPHNet \cite{poulenard2019effective} & 87.7 & 86.6 & 87.6 \\
    ClusterNet \cite{chen2019clusternet} & 87.1 & 87.1 & 87.1 \\
    GC-Conv \cite{zhang2020global} & 89.0 & 89.1 & 89.2 \\
    RI-Framework \cite{li2020rotation} & 89.4 & 89.4 & 89.3 \\
    \midrule
    \multicolumn{4}{c}{Ours} \\
    \midrule
    CN(frozen)-PointNet & 78.9 $\pm$ 2.1 & 78.7 $\pm$ 2.2 & 78.4 $\pm$ 2.5\\
    CN(L)-PointNet & 79.8 $\pm$ 1.4 & 79.6 $\pm$ 1.3 & 79.6 $\pm$ 1.4 \\
    CN(NL)-PointNet & 79.9 $\pm$ 1.3 & 79.6 $\pm$ 1.3 & 79.7 $\pm$ 1.3\\
    CN(frozen)-DGCNN & 88.3 $\pm$ 2.1 & 88.3 $\pm$ 2.1 & 88.3 $\pm$ 2.1\\
    CN(L)-DGCNN & 88.9 $\pm$ 1.8 & 88.6 $\pm$ 1.9 & 88.6 $\pm$ 2.0\\
    CN(NL)-DGCNN & 88.7 $\pm$ 1.8 & 88.8 $\pm$ 1.9 & 90.0 $\pm$ 1.1\\
    \bottomrule
  \end{tabular}
  }
  \vspace{-2mm}
\end{table}

We design our Canonicalization Network (CN) using layers from Vector Neurons \cite{deng2021vector}, where the final output contains three 3D vectors that are obtained by pooling over the entire point cloud. We then orthonormalize the three vectors using the Gram-Schmidt orthonormalization process to define a 3D ortho-normal coordinate frame or a rotation matrix. We canonicalize the point cloud by acting on it using this rotation matrix. We use a two-layered Vector Neuron network followed by global pooling, which we call CN(NL). To support our hypothesis that the canonicalization function can be inexpressive, we use a single linear layer of Vector neuron followed by pooling and call this model CN(L). Furthermore, to understand the significance of learning canonicalization, we freeze the weights of the CN and call this variant CN(frozen). We use PointNet and DGCNN \citep{wang2019dynamic} as the prediction networks in our experiments. 

\vspace{-1ex}
\paragraph{Results}
\cref{tab:exp:cls} contains the results of the ShapeNet experiment, showing the classification accuracy for different augmentation strategies during training and evaluation: $z/z$, $z/\mathrm{SO}(3)$, and $\mathrm{SO}(3)/\mathrm{SO}(3)$. Our method, which includes CN(frozen)-PointNet, CN(L)-PointNet, CN(NL)-PointNet, CN(frozen)-DGCNN, CN(L)-DGCNN, and CN(NL)-DGCNN, demonstrates competitive results across all rotation types. We achieve similar results in the ShapeNet part segmentation task as presented in \cref{tab:exp:seg}. In particular, we observe three trends in our results. First, learning canonicalization slightly improves the performance, except in the case where the test point clouds are already aligned ($z/z$ column of \cref{tab:exp:cls}). Second, using shallow linear canonicalization achieves good results. Third, the performance of the prediction network bottlenecks the model's performance. This verifies our hypothesis that decoupling the equivariance using a simple canonicalization network results in a better and more expressive non-equivariant prediction network to improve the performance of the task while still being equivariant. In \cref{tab:exp:inf_time}, we also show that the inference time of our algorithm is dominated by the prediction network's inference time.
The overhead of canonicalization is negligible, which makes our method faster than existing methods that modify the entire architecture like Vector Neurons \cite{deng2021vector}.
% \newcolumntype{R}{>{\columncolor{LightRed}}c}

% \newcolumntype{R}{>{\columncolor{LightRed}}c}
\begin{table}[th]
  \small
  \centering
  \caption{\small ShapeNet part segmentation results.  Overall average category mean IoU over 16 categories in two train/test scenarios are reported. $z$ here stands for aligned data augmented by random rotations around the vertical axis, and $\mathrm{SO}(3)$ indicates data augmented by random 3D rotations}
  \label{tab:exp:seg}
  \begin{tabular}{ccc}
    \toprule
    Methods & $z/\mathrm{SO}(3)$ & $\mathrm{SO}(3)/\mathrm{SO}(3)$ \\
    \midrule
    \multicolumn{3}{c}{Point / mesh inputs} \\
    \midrule
    PointNet \cite{qi2017pointnet} & 38.0 & 62.3 \\
    DGCNN \cite{wang2019dynamic} & 49.3 & 78.6 \\
    VN-PointNet\cite{deng2021vector} & 72.4 & 72.8 \\
    VN-DGCNN\cite{deng2021vector} & \textbf{81.4} & \textbf{81.4}\\
    PointCNN \cite{li2018pointcnn} & 34.7 & 71.4 \\
    PointNet++ \cite{qi2017pointnetpp} & 48.3 & 76.7 \\
    ShellNet \cite{zhang2019shellnet} & 47.2 & 77.1 \\
    RI-Conv \cite{zhang2019rotation} & 75.3 & 75.3 \\
    TFN \cite{thomas2018tensor} & 76.8 & 76.2 \\
    GC-Conv \cite{zhang2020global} & 77.2 & 77.3 \\
    RI-Framework \cite{li2020rotation} & 79.2 & 79.4 \\
    % --------------------
    % input: point + normal
    % --------------------
    \midrule
    \multicolumn{3}{c}{Ours} \\
    \midrule
    CN(frozen)-PointNet & 72.1 $\pm$ 0.8 & 72.3 $\pm$ 1.1\\
    CN(L)-PointNet & 73.4 $\pm$ 1.2 & 73.2 $\pm$ 0.9 \\
    CN(NL)-PointNet & 73.5 $\pm$ 0.8 & 73.6 $\pm$ 1.1 \\
    CN(frozen)-DGCNN & 78.1 $\pm$ 1.2 & 78.2 $\pm$ 1.2 \\
    CN(L)-DGCNN  & 78.5 $\pm$ 1.1 & 78.3 $\pm$ 1.2 \\
    CN(NL)-DGCNN & 78.4 $\pm$ 1.0 & 78.5 $\pm$ 0.9 \\
    \bottomrule
  \end{tabular}
  \vspace{-2mm}
\end{table}

\begin{table}[ht]
\centering
\caption{Inference time (in seconds) of the networks for ModelNet40 classification test split in 1 A100 and 8 CPUs with a batch size of 32. Vanilla denotes no modification to the base network, while Vector Neuron and Canonicalization denote that the base network is redesigned/enhanced with them to be equivariant.}
\label{tab:exp:inf_time}
\resizebox{\columnwidth}{!}{
\begin{tabular}{lccc}
\hline
Base Network & Vanilla & Vector Neuron & Canonicalization \\
\hline
PointNet & 18s & 30s & 20s\\
\hline
DGCNN & 23s &  39s & 25s\\
\hline
\end{tabular}}
\vspace{-2mm}
\end{table}

% \vspace{-1ex}
\subsection{Discussion}

Experimental results have shown the usefulness of our proposed method across different domains: images, $N$-body physical systems and point clouds. This supports the hypothesis that learning a canonicalization tends to perform better than using predefined heuristics to define it. This could be due to a combination of two factors. First, the learned canonicalizations have some consistency and help the prediction network perform the task. This is shown explicitly for our results in the image domain. Second, the process of learning the canonicalization induces an implicit augmentation of the data. This should help the prediction function generalize better and be more robust to potential failings of the canonicalization function. The method therefore combines some of the advantages of data augmentation with exact equivariance.

% \subsection{Image colorization}
% \subsubsection{Experimental setup}

% \subsubsection{Results}
% \vspace{-1ex}
\section{Conclusion}

In this work, we propose using a learned canonicalization function to obtain equivariant machine-learning models. These canonicalization functions can conveniently be plugged into existing architectures, resulting in highly expressive models. We have described general approaches to obtain canonicalization functions and specific implementation strategies for the Euclidean group (for images and point clouds) and the symmetric group.

We performed experimental studies in the image, dynamical systems and point cloud domains to test our hypotheses. First, we show that our approach achieves comparable or better performance than baselines on invariant tasks. Importantly, learning the canonical network is a better approach than using a fixed mapping, either a frozen neural network or a heuristic approach.
%Despite the fact that these alternatives also result in equivariant models, we think that learning the canonicalization function makes it less prone to discontinuities and facilitates the job of the prediction function.
Our results also show that the canonicalization function can be realized with a shallow equivariant network without hindering performance. Finally, we show that this approach reduces inference time and is more suitable for bigger groups than G-CNNs on images.

One limitation of our method is that there are no guarantees that the canonicalization function is smooth. This may be detrimental to generalization as small changes in the input could lead to large variations in the input to the prediction function. Another limitation could arise in domains in which semantic content is lacking to identify a meaningful canonicalization, for example, some types of astronomical images or biological images.

Multiple extensions of this framework are possible. Future work could include experimentation on canonicalization for the symmetric group. Other ways to build canonicalization functions could also be investigated, such as using steerable networks for images. The function would output an orientation fibre that transforms by the irreducible representation of the special orthogonal group. Understanding how design choices for canonicalization functions (for example, the subgroup $K$) affect downstream performance could also be a fruitful research direction. Finally, making large pretrained architectures equivariant using this framework could be an exciting extension.

% \vspace{-1ex}
\section*{Acknowledgements}

We thank Vasco Portilheiro for having provided an important correction in the proof of Theorem 3.3. We also thank Erik J. Bekkers, Pim de Haan, Aristide Baratin, Guillaume Huguet, Sébastien Lachapelle and Miltiadis Kofinas for their valuable comments. This project is in part supported by the CIFAR AI chairs program and NSERC Discovery. S.-O. K.'s research is also supported by IVADO and the DeepMind Scholarship. Computational resources were provided by Mila and Compute Canada.


\bibliography{paper}
\bibliographystyle{icml2023}

%%%%%%%%%%%%%%%%%%%%%%%%%%%%%%%%%%%%%%%%%%%%%%%%%%%%%%%%%%%%%%%%%%%%%%%%%%%%%%%
%%%%%%%%%%%%%%%%%%%%%%%%%%%%%%%%%%%%%%%%%%%%%%%%%%%%%%%%%%%%%%%%%%%%%%%%%%%%%%%
% APPENDIX
%%%%%%%%%%%%%%%%%%%%%%%%%%%%%%%%%%%%%%%%%%%%%%%%%%%%%%%%%%%%%%%%%%%%%%%%%%%%%%%
%%%%%%%%%%%%%%%%%%%%%%%%%%%%%%%%%%%%%%%%%%%%%%%%%%%%%%%%%%%%%%%%%%%%%%%%%%%%%%%
\newpage
\appendix
\onecolumn
\addcontentsline{toc}{section}{Appendix} % Add the appendix text to the document TOC
{\tiny
\part{Appendix} % Start the appendix part
}
\parttoc % Insert the appendix TOC

\vspace{-1ex}
\section{Symmetric inputs and relaxed equivariance}
\label{apd:symmetric}
An input $\v{x} \in \set{X}$ is symmetric if its stabilizer subgroup $G_{\v{x}} = \cbrace{g \in G \mid \rho\pr{g}\v{x} = \v{x}}$ is non-trivial. In other words, symmetric inputs are fixed by multiple group elements.

Given any $g_1, g_2 \in G$, we have $\rho\pr{g_1} \v{x} = \rho\pr{g_2} \v{x}$ if and only if $g_1$ and $g_2$ are part of the same coset for the stabilizer, e.g. $g_1, g_2 \in gG_{\v{x}}$. This follows from the well-known relation between orbits and stabilizers.
%Therefore, symmetric inputs are always fixed by multiple group elements.

Symmetric inputs are problematic when using the standard definition of equivariance for the canonicalization function because for $g_1, g_2 \in gG_{\v{x}}$, we have
\begin{align}
& h\pr{\rho\pr{g_1} \v{x}} = h\pr{\rho\pr{g_2} \v{x}},\\
& \rho\pr{g_1} h\pr{\v{x}} = \rho\pr{g_2} h\pr{\v{x}}.
\end{align}
If $g_1\neq g_2$, there cannot exist a $h\pr{\v{x}}\in \rho\pr{G}$ such that the last equality is satisfied.

We define a relaxed version of equivariance to address this.

\begin{definition}[Relaxed equivariance]
\label{def:relaxed}
Given group representations $\rho : G\to \mathrm{GL}\pr{\set{X}}$ and $\rho' : G\to \mathrm{GL}\pr{\set{Y}}$, a function $h : \set{X}\to\set{Y}$ satisfies relaxed equivariance if $\forall g_1, \v{x} \in G \times X $ there exists a $g_2 \in g_1 G_{\v{x}}$ such that
\begin{align}
& h\pr{\rho\pr{g_1}\v{x}} = \rho'\pr{g_2} h\pr{\v{x}}.
\end{align}
\end{definition}

Note that relaxed invariance coincides with standard invariance. It is possible to build canonicalization functions that satisfy this condition as we show in the next appendix.

We also prove the following important result.
\begin{theorem}
\label{th:relaxed} 
A function $\gv{\phi}$ defined by \cref{eq:model} satisfies the relaxed equivariance condition if the canonicalization function $h$ satisfies the relaxed equivariance condition.
\end{theorem}

\begin{proof}
We start with: 
\begin{align}
& \gv{\phi}\pr{\rho\pr{g_1}\v{x}} = h'\pr{\rho\pr{g_1}\v{x}} \v{f}\pr{h\pr{\rho\pr{g_1}\v{x}}^{-1} \rho\pr{g_1}\v{x} }
\end{align}
Using the definition of relaxed equivariance \ref{def:relaxed}, we obtain the following, where $g_2 \in g_1 G_{\v{x}}$ :
\begin{align}
& \gv{\phi}\pr{\rho\pr{g_1}\v{x}} = \rho'\pr{g_2}h'\pr{\v{x}} \v{f}\pr{h\pr{\v{x}}^{-1}\rho\pr{g_2}^{-1} \rho\pr{g_1}\v{x} }\\
& \gv{\phi}\pr{\rho\pr{g_1}\v{x}} = \rho'\pr{g_2}h'\pr{\v{x}} \v{f}\pr{h\pr{\v{x}}^{-1} \rho\pr{{g_2}^{-1}g_1}\v{x} }
\end{align}
Using the fact that ${g_2}^{-1}g_1 \in G_{\v{x}}$,
\begin{align}
& \gv{\phi}\pr{\rho\pr{g_1}\v{x}} = \rho'\pr{g_2}h'\pr{\v{x}} \v{f}\pr{h\pr{\v{x}}^{-1} \v{x} }\\
& \gv{\phi}\pr{\rho\pr{g_1}\v{x}} = \rho'\pr{g_2}\gv{\phi}\pr{\v{x}}
\end{align}
Therefore, $\gv{\phi}$ satisfies relaxed equivariance and this completes the proof.
\end{proof}

We provide more discussion on relaxed equivariance.
Standard equivariance requires that symmetric inputs are mapped to symmetric inputs. This is a limitation in many situations others than canonicalization that relxed equivariance allows to bypass.
% This is the same problem as we highlighted for canonicalization. If $\gv{\phi}$ is an equivariant function and $g_2 \in g_1 G_{\v{x}}$, it follows that
% \begin{align}
% \gv{\phi}\pr{\rho\pr{g_1} \v{x}} = \gv{\phi}\pr{\rho\pr{g_2} \v{x}},\\
% \rho\pr{g_1}  \gv{\phi}\pr{\v{x}} = \rho\pr{g_2} \gv{\phi}\pr{\v{x}},\label{eq:symmetry}
% \end{align}
% which means that $G_{{\phi}\pr{\v{x}}} \geq G_{\v{x}}$.
% The same is not true for relaxed equivariance since we can have 
% \begin{align}
% \gv{\phi}\pr{\rho\pr{g_2} \v{x}} = \rho\pr{g_1} \gv{\phi}\pr{\v{x}},
% \end{align}
% which removes the constraint of \cref{eq:symmetry} imposing symmetry of the output.

Aside from its importance for canonicalization, relaxed equivariance is thus a very useful concept in itself. It generalizes the idea of \textit{multiset-equivariance} from \citet{zhang2022multisetequivariant} to arbitrary groups. The relaxation captures the idea of equivariance up to symmetry and solves the inability of equivariant functions to break symmetry \cite{smidt2021,satorras2021n}. It should therefore be \textit{}{more} desirable than standard equivariance in many instances. For example, in physics, it allows to model symmetry breaking phenomenons or in graph representation learning, it allows to map nodes that are part of the same orbit to different embeddings.

The fact that we obtain relaxed equivariance for the canonicalized function is therefore a feature rather than a bug. This captures the desideratum that a function should be able to output asymmetric outputs from symmetric inputs.

\begin{figure}[h!]%{r}{0.5\linewidth}
\centering
    \begin{subfigure}[t]{0.45\textwidth}
        \centering
   \includegraphics[width=0.9\linewidth]{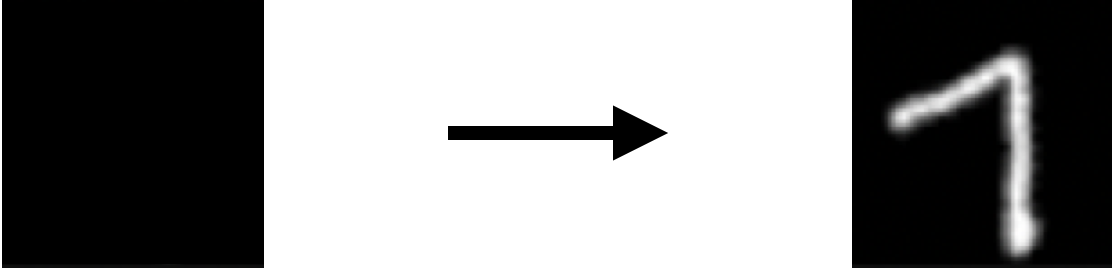}
        \caption{Translation or rotation equivariant function on images}
        \label{fig:sym_images}
    \end{subfigure}%
    \hfill
    \begin{subfigure}[t]{0.45\textwidth}
        \centering
   \includegraphics[width=0.9\linewidth]{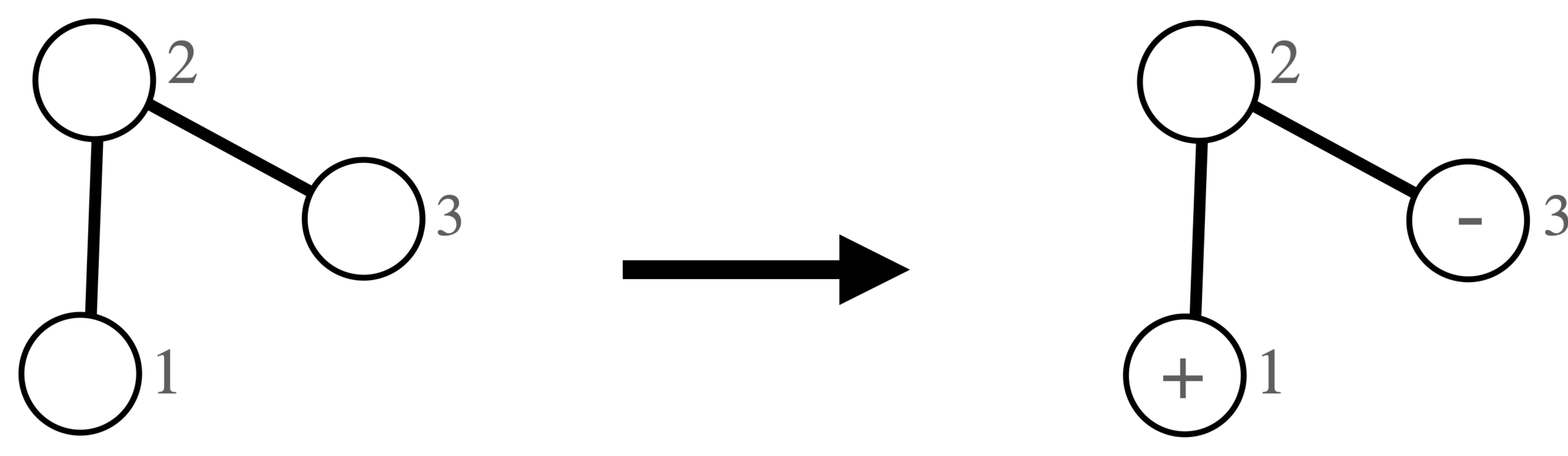}
        \caption{Permutation equivariant function on graphs}
        \label{fig:sym_graph}
    \end{subfigure}%
\caption{\small Example of task that cannot be performed by equivariant functions.}
  \label{fig:sym}
  \vspace{-2mm}
\end{figure}

Figure \ref{fig:sym} shows simple examples of function that cannot be approximated by equivariant functions. In \cref{fig:sym_images} assuming that the input is an image with uniform pixel values (fully symmetric), a translation equivariant function cannot output any image with non-uniform pixel values. In \cref{fig:sym_graph}, the input could be the symmetric graph of a molecule, with nodes $1$ and $3$ part of the same orbit. It is impossible for a permutation equivariant function to give different outputs (for example charge polarization) for these nodes. Yet, it is clear that in many situations such functions should not be excluded.

\section{Optimization approach to canonicalization}
\label{apd:optim}

In this appendix, we provide a more formal description of canonicalization functions obtained with the optimization approach of section \ref{sec:design}.

We prove a theorem providing a sufficient condition for a canonicalization function to satisfy the relaxed equivariance condition \ref{def:relaxed}.

\begin{theorem}
\label{th:canon}
Let $h\pr{\v{x}} \in\argmin_{\rho\pr{g}\in \rho\pr{G}} s\pr{ \rho\pr{g}, \v{x}}$ for some $s: \rho\pr{G} \times \set{X}\to \mathbb{R}$. If the conditions
\begin{enumerate}
    \item \label{cond:1} $ \forall g, g_1 \in G, \forall \v{x} \in \set{X}, \ s\pr{ \rho\pr{g}, \rho\pr{g_1}\v{x}} = s\pr{ \rho\pr{g_1}^{-1}\rho\pr{g}, \v{x}}$
    \item \label{cond:2} $\forall \v{x}, \exists g_1\in G$, such that $\argmin_{\rho\pr{g}\in \rho\pr{G}} s\pr{ \rho\pr{g}, \v{x}} \subseteq \rho\pr{G_{\v{x}}g_1}$ where $G_{\v{x}}$ is the stabilizer subgroup of $\v{x}$ 
\end{enumerate}
are satisfied then $h\pr{\v{x}}$ satisfies the relaxed equivariance condition \ref{def:relaxed}.
\end{theorem}
\begin{proof}
Let us introduce the shorthand notation $ \rho\pr{H_{\v{x}}} = \argmin_{\rho\pr{g}\in \rho\pr{G}} s\pr{ \rho\pr{g}, \v{x}}$. We define $H_{\v{x}}$ as a subset of $G$ such that its image under $\rho$ is the argmin.

We have
\begin{align}
\rho\pr{H_{\rho\pr{g_1}\v{x}}} = \argmin_{\rho\pr{g}\in \rho\pr{G}} s\pr{ \rho\pr{g}, \rho\pr{g_1}\v{x}}
\end{align}
Using condition \ref{cond:1}, we have
\begin{align}
\rho\pr{H_{\rho\pr{g_1}\v{x}}} =  \argmin_{\rho\pr{g}\in \rho\pr{G}} s\pr{ \rho\pr{g_1}^{-1}\rho\pr{g}, \v{x}}
\end{align}

We can use the fact that left multiplication by $\rho\pr{g_1}$ of the elements of $\rho\pr{H_{\v{x}}}$ will give the argmin in the previous equation. Therefore we have $\rho\pr{H_{\rho\pr{g_1}\v{x}}} = \rho\pr{g_1H_{\v{x}}}$.

Next, using condition \ref{cond:2}, there exists a $g_3\in G$ such that $\rho\pr{H_{\v{x}}} \subseteq \rho\pr{G_{\v{x}}g_3}$ and $\rho\pr{g_1H_{\v{x}}} \subseteq \rho\pr{g_1G_{\v{x}}g_3}$.

Finally, we can show that for any $h\pr{\v{x}} \in \rho\pr{G_{\v{x}}g_3}$ and $h\pr{\rho\pr{g_1}\v{x}} \in \rho\pr{g_1G_{\v{x}}g_3}$, there is a $g_2\in g_1G_{\v{x}}$ such that
\begin{align}
h\pr{\rho\pr{g_1}\v{x}} = \rho\pr{g_2} h\pr{\v{x}}
\end{align}
The left-hand side can be expressed as $h\pr{\rho\pr{g_1}\v{x}} = \rho\pr{g_1} h'\pr{\v{x}}$, where $h'\pr{\v{x}} \in \rho\pr{G_{\v{x}}g_3}$.
We then find
\begin{align}
\rho\pr{g_2} = \rho\pr{g_1} h'\pr{\v{x}}h\pr{\v{x}}^{-1}
\end{align}
Since $h\pr{\v{x}}$ and $h'\pr{\v{x}}$ are part of the same coset of the stabilizer, $h'\pr{\v{x}}h\pr{\v{x}}^{-1}$ must be part of the stabilizer. This completes the proof.

\end{proof}

Now let us discuss how the conditions of Theorem \ref{th:canon} can be met.

One way to satisfy the first condition is by using an equivariant function. Notice that the function $s: \rho\pr{G} \times \mathcal{X} \to \mathbb{R}$ can be reinterpreted as $s: \mathcal{X} \to \mathbb{R}^{ \rho\pr{G}}$. Therefore $s$ can be seen as a function of the input outputting a vector for which the components index the group representation. This vector should transform equivariantly for condition 1 to be satisfied, \emph{e.g.} $s\pr{\rho\pr{g_1}\v{x}}_{\rho\pr{g}} = s\pr{\v{x}}_{\rho\pr{g}_1^{-1}\rho\pr{g}}$.

Another way to satisfy condition 1 is to define $s\pr{ \rho\pr{g}, \v{x}} = u\pr{ \rho\pr{g}^{-1} \v{x}}$. It is easy to verify that this will indeed satisfy the condition.

Finally, as stated in the main text, condition 2 amounts to having a unique minimum in each orbit up an element of the stabilizer of the input. We will not show formally how this can be satisfied, but this is not expected to be a problem in practice. We conjecture that under weak assumptions, following the result of \citep{cox2020}, neural network functions can be obtained such that this condition is satisfied almost surely. In addition, for continuous groups, optimization can be made easier by making these neural network functions convex, which can be done using the framework of ICNN \citep{amos2017}.

\section{Proof of Theorem \ref{th:subgroup}}\label{apd:second}

We prove Theorem \ref{th:subgroup} which shows equivariance for a general subgroup $K$.

\begin{proof}
We have
\begin{align}
\gv{\phi}\pr{\rho\pr{g} \v{x}} = h'\pr{\rho\pr{g}\v{x}} \v{f}\pr{h\pr{\rho\pr{g}\v{x}}^{-1}\rho\pr{g}\v{x} }
\end{align}
If equation \ref{eq:general} is satisfied, then $\forall \ g, \v{x} \in G \times \set{X}$ there is a $k\in K$ such that
\begin{align}
\gv{\phi}\pr{\rho\pr{g} \v{x}} &= \rho'\pr{g} h'\pr{\v{x}} \rho'\pr{k} \v{f}\pr{\br{\rho\pr{g} h\pr{\v{x}} \rho\pr{k}^{-1}}^{-1} \rho\pr{g} \v{x} } \\
\gv{\phi}\pr{\rho\pr{g} \v{x}} &= \rho'\pr{g} h'\pr{\v{x}} \rho'\pr{k} \v{f}\pr{{\rho\pr{k}^{-1} h\pr{\v{x}}^{-1} \rho\pr{g}^{-1}} \rho\pr{g} \v{x} } 
\end{align}
Using the $K$-equivariance of $\v{f}$, we obtain
\begin{align}
\gv{\phi}\pr{\rho\pr{g} \v{x}} &= \rho'\pr{g} h'\pr{\v{x}} \rho'\pr{k} \rho'\pr{k}^{-1} \v{f}\pr{{ h\pr{\v{x}}^{-1} } \v{x} } \\
\gv{\phi}\pr{\rho\pr{g} \v{x}} &= \rho'\pr{g} h'\pr{\v{x}} \v{f}\pr{{ h\pr{\v{x}}^{-1} } \v{x} } 
\end{align}
\end{proof}

\section{Proof of Theorem \ref{th:normal}}\label{apd:normal}
\begin{proof}
We consider the special case where $K$ is a normal subgroup of $G$ such that the group can be taken to be isomorphic to a semidirect product $G \simeq K \rtimes J$. Then, group elements can be written as $g = \pr{k, j}$, where $k\in K$ and $j\in J$. The product is defined as $g_1 g_2 = \pr{k_1, j_1}\pr{k_2, j_2} = \pr{k_1 \varphi\br{j_1}\pr{k_2}, j_1 j_2}$, where $\varphi : J \to \mathrm{Aut}\pr{K}$ is a group homomorphism. Setting $k_2 = e$ and $j_1 = e$, we get any group element as $\pr{k_1, e}\pr{e, j_2} = \pr{k_1, j_2}$.

If the canonicalization function is $J$-equivariant and $K$-invariant, we have
\begin{align}
h\pr{\rho\pr{k,j}\v{x}} &= h\pr{\rho\pr{k,e}\rho\pr{e,j}\v{x}}\\
h\pr{\rho\pr{k,j}\v{x}} &= \rho\pr{e,j}h\pr{\v{x}}
\end{align}
We then show that there is a $k'\in K$ such that equation \ref{eq:general} is satisfied. Multiplying by $\rho\pr{e} = \rho\pr{k,e}\rho\pr{e,j} h\pr{\v{x}} h\pr{\v{x}}^{-1}\rho\pr{e,j}^{-1} \rho\pr{k,e}^{-1}$ on the left, we have
\begin{align}
\rho\pr{e,j}h\pr{\v{x}} &= \rho\pr{k,e}\rho\pr{e,j} h\pr{\v{x}} h\pr{\v{x}}^{-1}\rho\pr{e,j}^{-1} \rho\pr{k,e}^{-1}\rho\pr{e,j} h\pr{\v{x}}
\end{align}
Using the fact that conjugation of an element of $K$ by an element of $G$ preserves $K$ membership, we define $\rho\pr{k', e} = h\pr{\v{x}}^{-1}\rho\pr{e,j}^{-1} \rho\pr{k,e}^{-1}\rho\pr{e,j} h\pr{\v{x}}$
\begin{align}
\rho\pr{e,j}h\pr{\v{x}} &= \rho\pr{k,e}\rho\pr{e,j} h\pr{\v{x}} \rho\pr{k', e}
\end{align}
which shows that equation \ref{eq:general} is satisfied.
% \begin{align}
% \rho\pr{e,j}h\pr{\v{x}} &= \rho\pr{k,e}\rho\pr{e,j} h\pr{\v{x}} h\pr{\v{x}}^{-1} \rho\pr{k'',e}^{-1} h\pr{\v{x}}\\
% \rho\pr{e,j}h\pr{\v{x}} &= \rho\pr{k,e}\rho\pr{e,j} \rho\pr{k'',e}^{-1} h\pr{\v{x}}\\
% \rho\pr{e,j}h\pr{\v{x}} &= \rho\pr{k,e}\rho\pr{e,j}\rho\pr{e,j}^{-1} \rho\pr{k''',e}^{-1}\rho\pr{e,j} h\pr{\v{x}}\\
% \rho\pr{e,j}h\pr{\v{x}} &= \rho\pr{k,e}\rho\pr{k''',e}^{-1}\rho\pr{e,j} h\pr{\v{x}}
% \end{align}
% equality is satisfied for $k = k'''$.

Finally, we show that in this case, the image of $h$ can be chosen to be $\rho\pr{J}$. We first remark that in each orbit $\set{X}/G$ of the group action, the canonical sample $\hat{\v{x}}$ can be obtained from any orbit member $\v{x}$, as $\hat{\v{x}} = h\pr{\v{x}}^{-1} \v{x}$. For the canonical sample, we must have a $k\in K$ such that
\begin{align}
& h\pr{h\pr{\v{x}}^{-1} \v{x}} = h\pr{\v{x}}^{-1}  h\pr{\v{x}} \rho\pr{k,e}
\end{align}
If we impose $k = e$ to satisfy this condition, we have $h\pr{\hat{\v{x}}} = \rho\pr{e,e}$.

Since any orbit member can conversely be written as $\v{x} = \rho\pr{k,j} \hat{\v{x}}$ for some $k\in K$ and $j\in J$, if the canonicalization function is $J$-equivariant and $K$-invariant, we have
\begin{align}
h\pr{\v{x}} &= h\pr{\rho\pr{k,j} \hat{\v{x}}}\\
h\pr{\v{x}} &= \rho\pr{e,j} h\pr{\hat{\v{x}}}\\
h\pr{\v{x}} &= \rho\pr{e,j}
\end{align}
which completes the proof.    
\end{proof}

% \section{Lemma \ref{thm:compact}}

% \begin{lemma}\label{thm:compact}
% Let $\mathcal{K}\subseteq \set{X}$ be a compact subset of a finite-dimensional vector space $\set{X}$ and $f: \set{X}\to \mathrm{GL}\pr{\set{X}}$ be a continuous function. Then, the set $\widetilde{\mathcal{K}} = \cbrace{f\pr{\v{x}}\v{x}: \v{x}\in \mathcal{K}}$ is also compact.
% \end{lemma}

% \begin{proof}
% To show this, we need to show that the map $\v{x}\mapsto f\pr{\v{x}}\v{x}$ is continuous. This follows from the fact that it is the composition of the continuous maps $\v{x}\mapsto \pr{f\pr{\v{x}},\v{x}}$ and evaluation map $\pr{L, \v{x}} \mapsto L \v{x}$. The latter is continuous because $\set{X}$ is locally compact and Hausdorff.
% \end{proof}

\section{Proof of Theorem \ref{th:univ}}\label{apd:first}

\begin{proof}
The proof is inspired by the symmetrization approach of \cite{yarotsky2022universal} and \cite{puny2022frame}.

We first claim that given a compact set $\mathcal{K}\subseteq \set{X}$, the set $\widetilde{\mathcal{K}} = \cbrace{h\pr{\v{x}}^{-1}\v{x}: \v{x}\in \mathcal{K}}$ is also compact.

To see this, we notice that the map $\v{x} \mapsto h\pr{\v{x}}^{-1}$ is continuous, since it is the composition of the continuous function $h$ and of the inverse map $L \mapsto L^{-1}$. We use the fact that linear operators on $\set{X}$ form a Banach algebra and that the inverse map on Banach algebras is continuous. The map $\v{x} \mapsto h\pr{\v{x}}^{-1}$ is then composed with the evaluation map $\pr{L, \v{x}} \mapsto L \v{x}$. The latter is continuous since $\set{X}$ is locally compact and Hausdorff.

Then, let $\gv{\psi}$ be an arbitrary $G$-equivariant function, and $\gv{\phi}$ be defined by equation \ref{eq:model}. We have
\begin{align}
\norm{\gv{\psi}\pr{\v{x}} - \gv{\phi}\pr{\v{x} }} &= \norm{\gv{\psi}\pr{\v{x}} - h'\pr{\v{x}}  \v{f}\pr{h\pr{\v{x}}^{-1} \v{x} }}.
\end{align}

By the equivariance of $\gv{\psi}$, we obtain
\begin{align}
\norm{\gv{\psi}\pr{\v{x}} - \gv{\phi}\pr{\v{x} }} &= \norm{h'\pr{\v{x}}  \gv{\psi}\pr{h\pr{\v{x}}^{-1} \v{x} } - h'\pr{\v{x}}  \v{f}\pr{h\pr{\v{x}}^{-1} \v{x} }}.
\end{align}

We have that $\norm{h'\pr{\v{x}}}$ is bounded on $\mathcal{K}$ from continuity of $h'$ and of the induced operator norm. We therefore define $c = \sup_{\v{x} \in \mathcal{K}} \norm{h'\pr{\v{x}}} > 0$ and obtain
\begin{align}
\norm{\gv{\psi}\pr{\v{x}} - \gv{\phi}\pr{\v{x} }} &\leq \norm{h\pr{\v{x}}} \norm{\gv{\psi}\pr{h\pr{\v{x}}^{-1} \v{x} } -  \v{f}\pr{h\pr{\v{x}}^{-1} \v{x} }},\\
\norm{\gv{\psi}\pr{\v{x}} - \gv{\phi}\pr{\v{x} }} &\leq c \norm{\gv{\psi}\pr{h\pr{\v{x}}^{-1} \v{x} } -  \v{f}\pr{h\pr{\v{x}}^{-1} \v{x} }}, \ \forall \ \v{x} \in \set{K}. \label{eq:ineq}
\end{align}

Using the assumption that $\v{f}$ is a universal approximator of $K$-equivariant functions, we know that it is also a universal approximator of $G$-equivariant functions. We therefore have for any $\delta > 0$,
\begin{align}
\norm{\gv{\psi}\pr{\tilde{\v{x}} } -  \v{f}\pr{\tilde{\v{x}} }} \leq \delta, \ \forall \ \tilde{\v{x}} \in \widetilde{\mathcal{K}}.
\end{align}
In particular, we consider $\delta = \epsilon / c$. Replacing in \cref{eq:ineq},  we obtain the desired result
\begin{align}
\norm{\gv{\psi}\pr{\v{x}} - \gv{\phi}\pr{\v{x} }} &\leq \epsilon, \ \forall \ \v{x} \in \set{K}.
\end{align}

\end{proof}

\section{Proof of Theorem 5}
\label{apd:gram}

\begin{theorem}
The Gram-Schmidt process is $O\pr{n}$-equivariant.
\end{theorem}

\begin{proof}
Given $n$ linearly independent input vectors $\v{v}_1, \dots, \v{v}_n$, the Gram-Schmidt process first produces the orthogonal vectors $\v{u}_1, \dots, \v{u}_n$, with
\begin{align}
\label{eq:proj}
& \v{u}_i \pr{\v{v}_1, \dots, \v{v}_n} = \v{v}_i - \sum_{j=1}^{i-1} \frac{\v{u}_j \cdot \v{v}_i}{\norm{\v{u}_j}^2} \v{u}_j
\end{align}
The orthonormal basis $\v{e}_1, \dots, \v{e}_n$ is then given by
\begin{align}
\label{eq:normalize}
& \v{e}_i \pr{\v{v}_1, \dots, \v{v}_n} = \frac{\v{u}_i}{\norm{\v{u}_i}}
\end{align}

We wish to prove that $\forall i \leq n, \v{O}\in O\pr{n}$, we have
\begin{align}
& \v{e}_i \pr{\v{O}\v{v}_1, \dots, \v{O}\v{v}_n} = \v{O} \v{e}_i \pr{\v{v}_1, \dots, \v{v}_n}
\end{align}

We first prove equivariance of \eqref{eq:proj} by strong induction. Consider the base case with $i=1$. We have $\v{u}_1\pr{\v{v}_1, \dots, \v{v}_n} = \v{v}_1$, which is trivially equivariant.
Then, we make the induction hypothesis that \eqref{eq:proj} is equivariant for $ 1 \leq i \leq k$. We can show that this implies equivariance for $i = k+1$.
We have
\begin{align}
\v{u}_{k+1}\pr{\v{v}_1, \dots, \v{v}_n} = \v{v}_{k+1} - \sum_{j=1}^{k} \frac{\v{u}_j \cdot \v{v}_{k+1}}{\norm{\v{u}_j}^2} \v{u}_j
\end{align}
Using the induction hypothesis, we obtain
\begin{align}
\v{u}_{k+1}\pr{\v{O}\v{v}_1, \dots, \v{O}\v{v}_n} = \v{O}\v{v}_{k+1} - \sum_{j=1}^{k} \frac{\v{O}\v{u}_j \cdot \v{O}\v{v}_{k+1}}{\norm{\v{O}\v{u}_j}^2} \v{O}\v{u}_j
\end{align}
Since the dot product and the Euclidean norm are $O\pr{n}$-invariant, we obtain
\begin{align}
\v{u}_{k+1}\pr{\v{O}\v{v}_1, \dots, \v{O}\v{v}_n} = \v{O}\v{v}_{k+1} - \sum_{j=1}^{k} \frac{\v{u}_j \cdot \v{v}_{k+1}}{\norm{\v{u}_j}^2} \v{O}\v{u}_j\\
\v{u}_{k+1}\pr{\v{O}\v{v}_1, \dots, \v{O}\v{v}_n} = \v{O}\pr{\v{v}_{k+1} - \sum_{j=1}^{k} \frac{\v{u}_j \cdot \v{v}_{k+1}}{\norm{\v{u}_j}^2} \v{u}_j}
\end{align}
which completes the induction.

We finally see that by $O\pr{n}$-invariance of the Euclidean norm, \eqref{eq:normalize} is also equivariant. Since the composition of equivariant functions is equivariant, we find that the Gram-Schmidt process is equivariant and this completes the proof.

\end{proof}

\section{Implementation details}

\subsection{Image classification experiments}
\label{appndx:image:exp}
\paragraph{Training details.} In all our image experiments, we train the models by minimizing the cross entropy loss for 100 epochs using Adam \citep{adam} with a learning rate of 0.001. We perform early stopping based on the classification performance of the validation dataset with a patience of 20 epochs.
 \paragraph{CNN architecture.} For CNN (base), we choose an architecture with 7 layers where layer 1 to 3 has 32, 4 to 6 has 64, and layer 7 has 128 channels, respectively. Instead of pooling, we use convolution filters of size $5\times 5$ with a stride 2 at layers 4 and 7. The remaining convolutions have filters of size $3\times 3$ and stride 1. All the layers are followed by batch-norm \cite{ioffe2015batch} and ReLU activation with dropout(p=0.4) only at layers 4 and 7. 
 \paragraph{G-CNN architecture.} We took the same CNN architecture as above and replaced the standard convolutions with group convolutions \cite{cohen2016group}.
\paragraph{Optimization approach.} For the energy function $E$, the image is transformed to a point cloud and fed into a Deep Sets \cite{zaheer2017deep} architecture. Then, $E$ is optimized by 5-steps of gradient descent (learning rate 0.1) using implicit differentiation.

\subsection{$N$-body dynamics prediction experiments}
\label{apd:n-body}

\paragraph{Training details.}
We train on mean square error (MSE) loss between predicted and ground truth using the Adam optimizer. We train for 10.000 epochs and use early stopping. We use weight decay $10^{-8}$ and dropout in the canonicalization function with $p=0.5$.

\paragraph{Canonicalization network architecture.} We use a Vector Neurons version of the Deep Sets architecture for the canonicalization network in this task. The network has two layers with hidden dimension size of 32.

\paragraph{Prediction network architecture.} The GNN prediction network uses the same architecture as \citep{satorras2021n}.

\subsection{Point Cloud Classification and Segmentation experiments}

\paragraph{Training Details} We use cross entropy loss and Stochastic Gradient Descent (SGD) optimizer to train the network for $200$ epochs in all of our pointcloud experiments. We use a initial learning rate of $0.1$ and cosine annealing schedule with an end learning rate of $0.001$.

\paragraph{Canonicalization network architecture.} We design our Canonicalization Network (CN) using layers from Vector Neurons \cite{deng2021vector}, where the final output contains three 3D vectors that are obtained by pooling over the entire point cloud. We then orthonormalize the three vectors using the Gram-Schmidt orthonormalization process to define a 3D ortho-normal coordinate frame or a rotation matrix.

\paragraph{Prediction network architecture.} We use PointNet and DGCNN \citep{wang2019dynamic} as the prediction networks in our experiments. 
% \newpage

\section{Additional results}

\subsection{Image classification}
\label{appdx:image:add_results}

\begin{figure*}[ht]
% \floatconts
  \centering
  \begin{subfigure}{0.15\textwidth}
      \includegraphics[trim={0cm 9.3cm 0cm 5.1cm}, clip, width=1.0\linewidth]{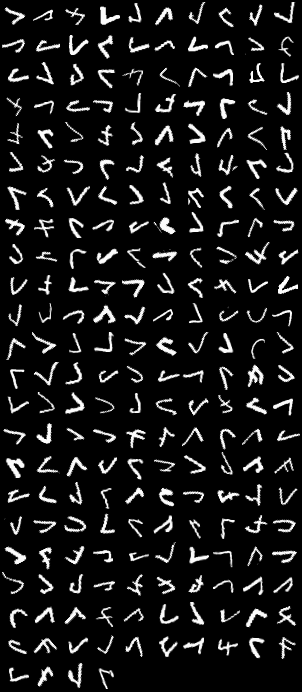}
      \caption{Original}
  \end{subfigure}\hfill
  \begin{subfigure}{0.15\textwidth}
      \includegraphics[trim={0cm 9.3cm 0cm 5.1cm}, clip, width=1.0\linewidth]{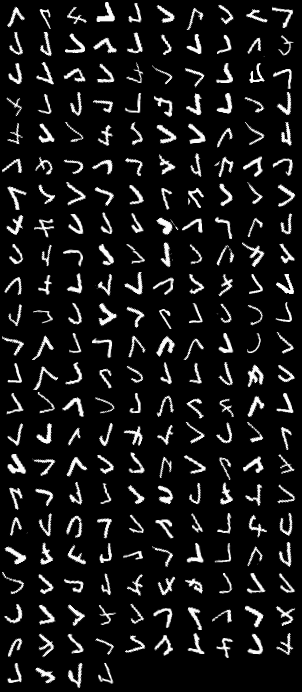}
      \caption{p4}
  \end{subfigure}\hfill
  \begin{subfigure}{0.15\textwidth}
      \includegraphics[trim={0cm 9.3cm 0cm 5.1cm}, clip, width=1.0\linewidth]{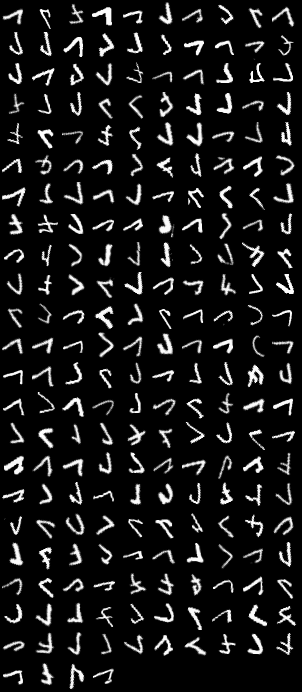}
      \caption{p8}
  \end{subfigure}\hfill
  \begin{subfigure}{0.15\textwidth}
      \includegraphics[trim={0cm 9.3cm 0cm 5.1cm}, clip, width=1.0\linewidth]{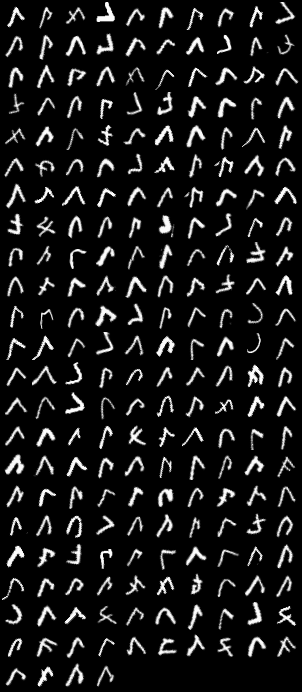}
      \caption{p16}
  \end{subfigure}\\

  \begin{subfigure}{0.15\textwidth}
      \includegraphics[trim={0cm 9.3cm 0cm 5.1cm}, clip, width=1.0\linewidth]{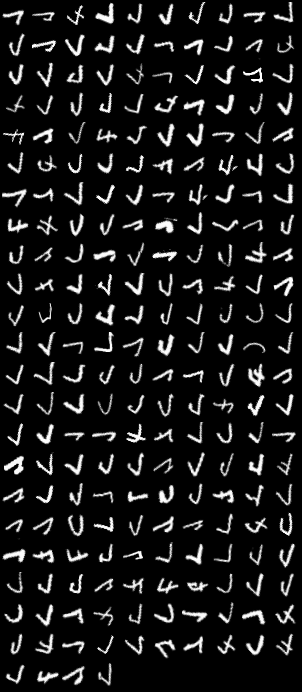}
      \caption{p32}
  \end{subfigure}\hfill
  \begin{subfigure}{0.15\textwidth}
      \includegraphics[trim={0cm 9.3cm 0cm 5.1cm}, clip, width=1.0\linewidth]{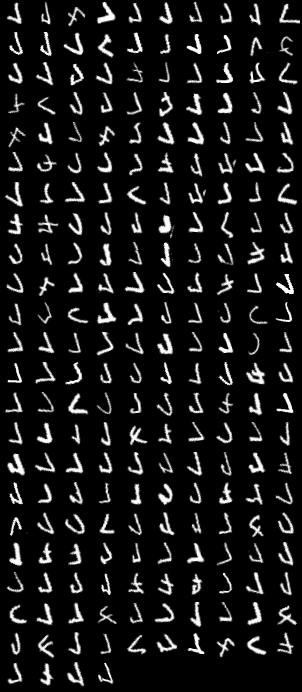}
      \caption{p64}
  \end{subfigure}\hfill
  \begin{subfigure}{0.15\textwidth}
      \includegraphics[trim={0cm 9.3cm 0cm 5.1cm}, clip, width=1.0\linewidth]{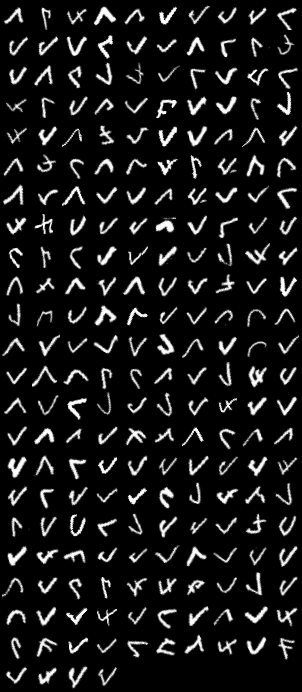}
      \caption{p64 (frozen)}
  \end{subfigure}\hfill
  \begin{subfigure}{0.15\textwidth}
      \includegraphics[trim={0cm 9.3cm 0cm 5.1cm}, clip, width=1.0\linewidth]{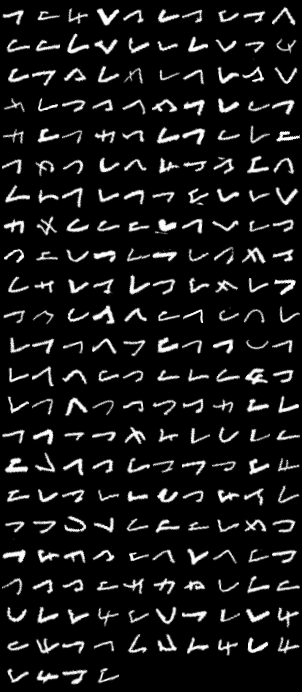}
      \caption{PCA}
  \end{subfigure}\\

% \floatconts
  % \label{fig:subfigex}
  % {\caption{Canonicalized images from different canonicalization functions for digit 1.}}
  \centering
  \begin{subfigure}{0.15\textwidth}
      \includegraphics[trim={0cm 9.3cm 0cm 5.1cm}, clip, width=1.0\linewidth]{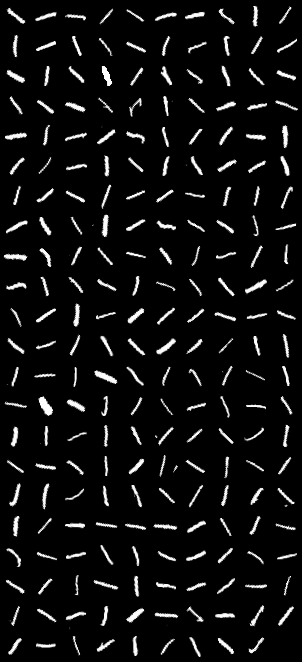}
      \caption{Original}
  \end{subfigure}\hfill
  \begin{subfigure}{0.15\textwidth}
      \includegraphics[trim={0cm 9.3cm 0cm 5.1cm}, clip, width=1.0\linewidth]{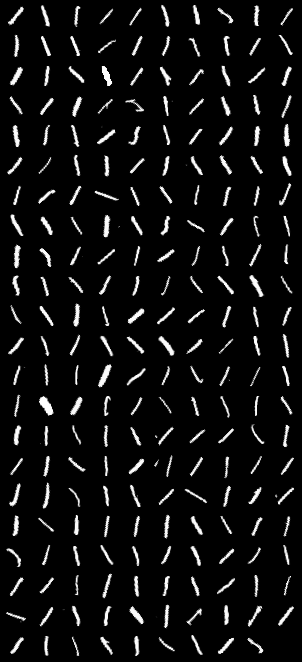}
      \caption{p4}
  \end{subfigure}\hfill
  \begin{subfigure}{0.15\textwidth}
      \includegraphics[trim={0cm 9.3cm 0cm 5.1cm}, clip, width=1.0\linewidth]{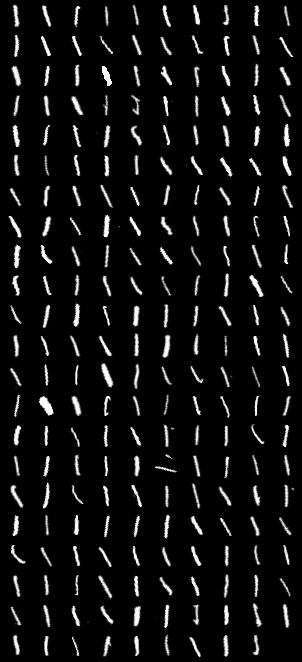}
      \caption{p8}
  \end{subfigure}\hfill
  \begin{subfigure}{0.15\textwidth}
      \includegraphics[trim={0cm 9.3cm 0cm 5.1cm}, clip, width=1.0\linewidth]{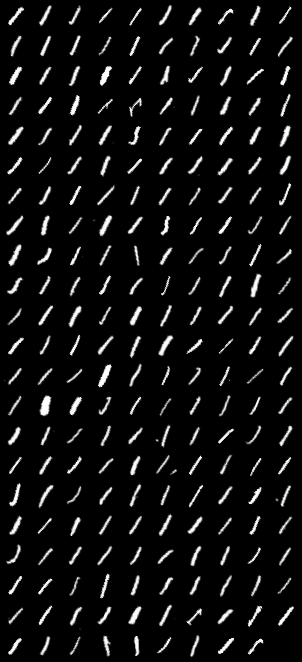}
      \caption{p16}
  \end{subfigure}\\

  \begin{subfigure}{0.15\textwidth}
      \includegraphics[trim={0cm 9.3cm 0cm 5.1cm}, clip, width=1.0\linewidth]{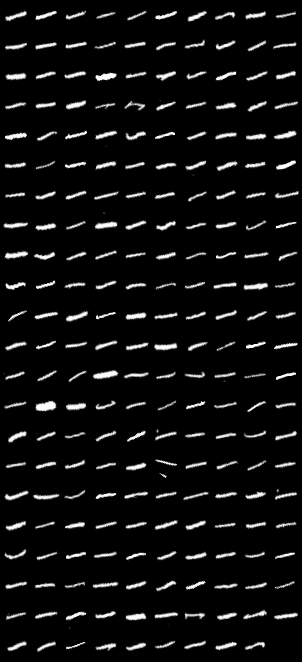}
      \caption{p32}
  \end{subfigure}\hfill
  \begin{subfigure}{0.15\textwidth}
      \includegraphics[trim={0cm 9.3cm 0cm 5.1cm}, clip, width=1.0\linewidth]{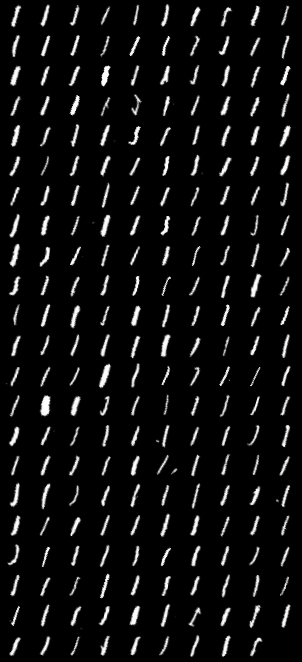}
      \caption{p64}
  \end{subfigure}\hfill
  \begin{subfigure}{0.15\textwidth}
      \includegraphics[trim={0cm 9.3cm 0cm 5.1cm}, clip, width=1.0\linewidth]{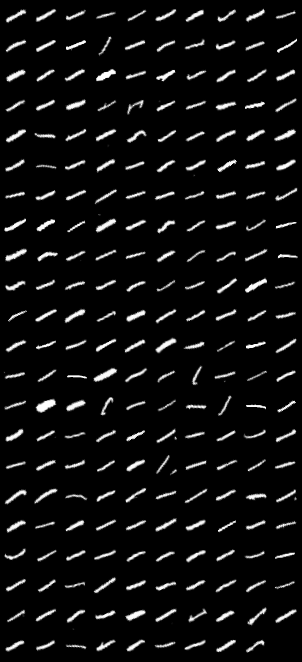}
      \caption{p64 (frozen)}
  \end{subfigure}\hfill
  \begin{subfigure}{0.15\textwidth}
      \includegraphics[trim={0cm 9.3cm 0cm 5.1cm}, clip, width=1.0\linewidth]{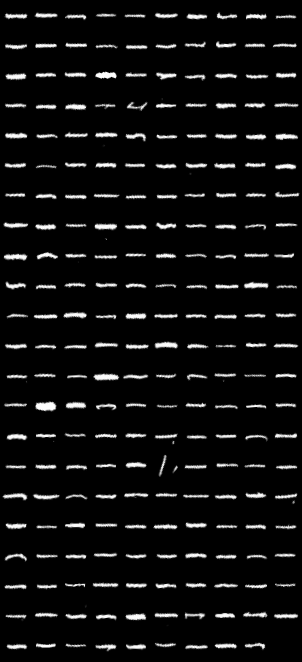}
      \caption{PCA}
  \end{subfigure}
  \caption{\small Canonicalized images from different canonicalization functions for digit 7.}
  \label{fig:canonized7}
\end{figure*}

\begin{table}[ht]
\small
\centering
{\caption{Impact of the number of layers in canonicalization function network and order of the discrete rotations to which it is equivariant on the performance.}}
\begin{adjustbox}{width=0.7\linewidth}
  {\begin{tabular}{c|cccccc}
  \toprule
  \bfseries  \#lyrs & \multicolumn{5}{c}{ \bfseries Order of the discrete rotation group} \\
    & p4 & p8 & p16 & p32 & p64& \\
  \midrule
  1 & 2.52 $\pm$ 0.12& 2.37 $\pm$ 0.09& 2.20 $\pm$ 0.08 & 2.05 $\pm$ 0.15 & 2.01 $\pm$ 0.09\\
  2 & 2.44 $\pm$ 0.06& 2.31 $\pm$ 0.05& 2.16 $\pm$ 0.09 & 2.00 $\pm$ 0.07 & 2.02 $\pm$ 0.12\\
  3 & 2.41 $\pm$ 0.11 & 2.28 $\pm$ 0.09& 2.11 $\pm$ 0.06 & 1.98 $\pm$ 0.09 & 1.99 $\pm$ 0.10\\
  \bottomrule
  \end{tabular}}
  \end{adjustbox}
\label{tab:ablation1}
\end{table}
First, we vary the number of layers of the canonicalization network and the number of rotations it is equivariant to. For this, we extend the layers of G-CNN to any arbitrary rotations. As we noticed that using a larger filter leads to better performance for higher order rotations, we stick to architecture with a lifting layer with image-sized filters followed by $1\times 1$ filters. From Table 6, we notice that adding equivariance to higher order rotation in the canonicalization function leads to significant performance improvement compared to adding more layers. Figure \ref{fig:canonized7} shows the canonical orientation resulting from the learnt canonicalization function with a single lifting layer on 90 randomly sampled images of class 7 from the test dataset. This suggests that a shallow network is sufficient to achieve good results with a sufficiently high order of discrete rotations. For p64, we see that all the similar-looking samples are aligned in one particular orientation. In contrast, although techniques like PCA or freezing parameters of the canonicalization function find the correct canonicalization function for simple digits like 1, they struggle to find stable mappings for more complicated digits like $7$. 

% \subsection{Visualizations of image canonicalizations}

\clearpage

\section{Algorithm for Image Inputs}
\label{apd:third}
\begin{algorithm}[htp]
    \caption{Differentiable Canonicalization for Image Inputs}
    \label{alg:alpha_model_selection}
    
     \definecolor{codeblue}{rgb}{0.25,0.5,0.5}
     \lstset{
       basicstyle=\fontsize{8pt}{8pt}\ttfamily\bfseries,
       commentstyle=\fontsize{8pt}{8pt}\color{codeblue},
       keywordstyle=\fontsize{8pt}{8pt},
     }
 \begin{lstlisting}[language=python, mathescape=true]
import torch.nn.functional as F
import kornia as K

def get_canonicalized_images(images, fibre_features, use_reflection=True):
    """
    images: Tensor with shape (batch_size, in_channels, height, width)
    fibres_features: Tensor with shape: (batch_size, num_group_elements)
    use_reflection: Boolean
    :return: (batch_size, in_channels, height, width)
    """
    num_group_elements = fibre_features.shape[-1]
    num_rotations = num_group_elements // 2 if use_reflection else num_group_elements
    
    fibre_features_one_hot = F.one_hot(
        torch.argmax(fibre_features, dim=-1), 
        num_group_elements
    ).float()
    
    fibre_features_soft = F.softmax(fibre_features, dim=-1)
    ref_angles = torch.linspace(0., 360., num_rotations+1)[:num_rotations]
    
    if use_reflection:
        ref_angles = torch.cat([ref_angles, ref_angles], dim=0)
        
    angles = torch.sum((
        fibre_features_one_hot + fibre_features_soft - fibre_features_soft.detach()
        ) * ref_angles, dim=-1)
        
    if use_reflection:
        reflect_one_hot = torch.cat(
            [torch.zeros(num_rotations), torch.ones(num_rotations)]
            , dim=0)
        reflect_indicator = torch.sum((
            fibre_features_one_hot + fibre_features_soft - fibre_features_soft.detach()
            ) * reflect_one_hot, dim=-1)
            
        images_reflected = K.geometry.hflip(images)
        reflect_indicator = reflect_indicator[:,None,None,None]
        images = (1 - reflect_indicator) * images + reflect_indicator * images_reflected
        
    return K.geometry.rotate(images, -angles)
    

# Use a shallow G-CNN as a canonicalization_network
feature_map = canonicalization_network(images)
# feature_map shape: (batch_size, num_channels, num_group_elements, height, width)

fibre_features = feature_map.mean(dim=(1, 3, 4))
# fibre_features shape: (batch_size, num_group_elements)

canonicalized_images = get_canonicalized_images(images, fibre_featuresr)

 \end{lstlisting}
 \label{alg:alpha_linear_eval}
 \end{algorithm}

%%%%%%%%%%%%%%%%%%%%%%%%%%%%%%%%%%%%%%%%%%%%%%%%%%%%%%%%%%%%%%%%%%%%%%%%%%%%%%%
%%%%%%%%%%%%%%%%%%%%%%%%%%%%%%%%%%%%%%%%%%%%%%%%%%%%%%%%%%%%%%%%%%%%%%%%%%%%%%%

\end{document}